# Clocks all the way down: does sequence-processing AI need endogenous temporal structure?

Meysam Hashemi[1#*], Alireza Valizadeh[2#], Viktor Jirsa[1]

[1]Institut de Neurosciences des Systèmes, Aix Marseille Université, INSERM, Marseille, 13005, France.

[2]Zapata Briceño Institute for Human Intelligence Research, Madrid, Spain.

meysam.hashemi@univ-amu.fr, av@institutoz-b.org, viktor.jirsa@univ-amu.fr

## Abstract

Mainstream sequence-processing AI is temporally reactive: each update is driven by input and learned state transitions, rather than by an independently evolving internal temporal reference. Biological brains, by contrast, continuously generate rhythmic activity that structures perception, memory, and cognition, and this rhythmicity reaches all the way to behavior. We ask whether this difference is merely implementational or genuinely computational. Four lines of evidence suggest that endogenous temporal structure supports putative functions that input-driven architectures may not readily replicate: 1) phase-dependent routing of information, 2) delay-structured combinatorial representation, 3) cross-frequency multiplexing of items, and 4) oscillation-gated offline consolidation. A proof-of-concept delay-ablation test dissociates representational capacity from memory retrieval. In recurrent and spatial spiking networks with fixed synaptic weights and connectivity, shuffling delays preserves part of the capacity but substantially impairs the sequence retrieval. The relative contrast is more pronounced for spatial networks and longer propagations. Instantaneous transmission collapses capacity itself. We then outline a cross-frequency multiplexing scheme that combines item identity and serial order with alternative mechanisms for selective memory retrieval. Finally, we state one falsifiable hypothesis, conditions for matched-resource tests, and three research directions: the components already exist; the disciplined comparison does not.

[#]These authors contributed equally. *Correspondence: meysam.hashemi@univ-amu.fr

## Highlights

• Mainstream sequence models treat time as input (order or timestamps), not as an internal temporal reference; brains instead use oscillations, delays, and phase to route, store, order, and consolidate information—on their own schedule.

• Perception and attention vary with the phase of ongoing rhythms even when input is aperiodic, so brains fail the timing-invariance test that reactive architectures pass by construction; whether this rhythmicity is an advantage remains open.

• Four functions reactive models do not obviously realize: flexible routing (via phase), high-capacity representation (via delays), ordered multi-item memory (nested rhythms within episodes), and offline consolidation (between episodes).

• Heterogeneous time delays expand the combinatorial representation beyond network size. Shuffling delays degrades memory retrieval far more than capacity, whereas removing them collapses capacity to zero in the networks tested.

## The missing clock

Ask a large language model what "time" it is, and it can answer readily—but only because current temporal information is supplied externally, for example through a timestamp, system context, or access to a system clock. The base model fundamentally lacks any internal sense of passing time. Sequence models [1] encode order through input-driven computational primitives such as positional indices or recurrent state transitions, but sequence position is not itself a measure of elapsed time, nor does it provide an independently generated temporal reference. In a standard transformer model, inference operates on the supplied context without maintaining a persistent state across calls or an intrinsic clock against which elapsed time can be measured. Instead, order enters through positional encodings (absolute, relative or rotary) [2, 3, 4] indexed by token position. Agents built around such a model can be given temporal cues, memory, and scheduled triggers [5], but these are external scaffolds: they allow the system to track time operationally without giving the model an intrinsic representation of time passing. Yet, such scaffolding does not fully close the gap: agents may still fail to recognize when stored information has become outdated, resulting in redundant calls or failure to execute necessary actions [6], and may miss real-time deadlines despite active monitoring [7]. For many tasks, atemporal processing is sufficient. But temporal cognition is different: perception unfolds over time (integrating successive inputs) [8], memory must be indexed by recency and sequence (when and in what order) [9], decisions depend on prospective coding (anticipating future states) [10], and communication relies on synchrony (coordinating events in time) [11]. None of these are naturally static-mapping problems. For example, speech comprehension requires the brain to maintain persistent, parallel, and time-stamped representations of a transient input stream, preserving content and relative temporal order after the original sensory events have passed [12].

Brains meet these demands through neural oscillations [13]: rhythmic activity ranging from slow delta waves (1–4 Hz) to fast gamma bursts (30–80 Hz), providing an "internal clock"; rather than a single centralized timekeeper, the brain constructs temporal reference frames through interacting neural oscillations across scales [14]. Within this framework, nested rhythms organize information flow, memory, and temporal prediction through cross-scale interactions, such as phase-amplitude coupling [15, 16, 17]. The idea that the temporal dynamics of neural activity—manifested as synchrony, phase lag, and precise spike timing—bind information across distributed units [18, 19], gate the selective routing of information [20], and convey top-down predictions [16] is long-standing in systems neuroscience. Whether sequence-processing AI needs an analogous mechanism, or whether existing architectures already approximate these functions through other primitives, remains an open question. We argue that internally generated temporal dynamics may offer computational advantages for specific functions over reactive, input-driven processing. This hypothesis hinges on how temporal structure enters computation: whether it is supplied externally through the input stream or arises internally as temporal variables and constraints that govern computation. We refer to the latter as ***endogenous temporal structure***: self-generated rhythms (e.g., oscillatory phase) and internal timing relations (e.g., transmission delays), as opposed to temporal information supplied externally to sequence models through input (Box 1).

**Box 1. What 'endogenous' means**

Two systems can both track sequence and yet differ in how time enters their computation: as an externally imposed input or an internally generated process (see Fig. 1).

**Exogenous, or reactive.** A *transformer* [2] receives an ordered list of input elements tagged with positional codes. Sequence order is externally specified rather than generated by the network's own dynamics. *State-space models* [21] maintain a summary of the past, a genuine memory, but each update is driven by the arrival of the next input. *Autoregressive generation* [2] does not change this classification, although it can appear to: the model can continue indefinitely by consuming its own output, yet each update is triggered by the next generation step, not by the passage of time. The operational test is timing invariance: absent externally supplied timestamps, a reactive system computes identically whether the physical intervals between successive inputs are milliseconds or hours, because its dynamics are indexed by arrival order, rather than elapsed duration. These systems represent sequence and history without generating their own temporal reference.

**Endogenous, or self-generated.** A cortical circuit can generate self-sustained oscillations even in the absence of structured input: in one canonical mechanism, excitation drives activity, delayed inhibition suppresses it, and as inhibition decays, the cycle repeats [22]. Such a circuit fails timing invariance by construction: its state advances between inputs, each arrival meets a different phase, and the interval itself becomes readable. *Transmission delays* impose stable phase lags between connected regions [23], so timing is organized by the circuit's own dynamics rather than by external input. In the hippocampus, cell-assembly sequences advance even when the sensory input is held constant, as when an animal runs in place, so that elapsed time is read from autonomous dynamics rather than from changing incoming input [24]. Time is therefore an internal variable that other computations can read and write, not a label attached to the input. Endogenous does not mean isolated: external events can reset or entrain the system's internal temporal structure, but do not create it: remove the external input and the circuit keeps cycling [14].

**Boundary cases.** Several recent architectures already blur the dichotomy. A class of deep neural networks known as *neural ordinary differential equations* [25] admits autonomous dynamics through continuous-time state evolution with adaptive step sizes and constant memory cost, but may drift in phase over long time horizons, especially for high-frequency components [26]. As in other continuous-time dynamical systems, discretization introduces temporal residual connections that stabilize gradient-based learning, while oscillations modulate gradients and provide an inductive bias that expands the dynamical repertoire to include resonance, enhancing performance relative to conventional networks. Inspired by cortical dynamics, configuring each node of a recurrent network as a damped harmonic oscillator (representing an *E–I* population) allows information encoding in amplitude and frequency bands, e.g., stimulus sequence order in phase shifts [27]. Implementing features such as heterogeneous frequencies and inhomogeneous delays, further enhanced the classification accuracy and learning speed. Oscillatory state-space models also outperform several alternatives in learning dynamical systems and long sequences [28]. Yet in current practice the temporal variable is externally supplied (an integration span, a discretization step), even when the modeled dynamics are autonomous. Conversely, endogenous oscillations need not be built in: recurrent networks trained on a working-memory task, given access to an external reference, learn two phase-coupled oscillators, one generated by the network and the other coupled to the external reference, and the phase difference between them maintains the identity of transient stimuli [29]. These systems mark the frontier where the two sides begin to blur, and fair tests must include them.

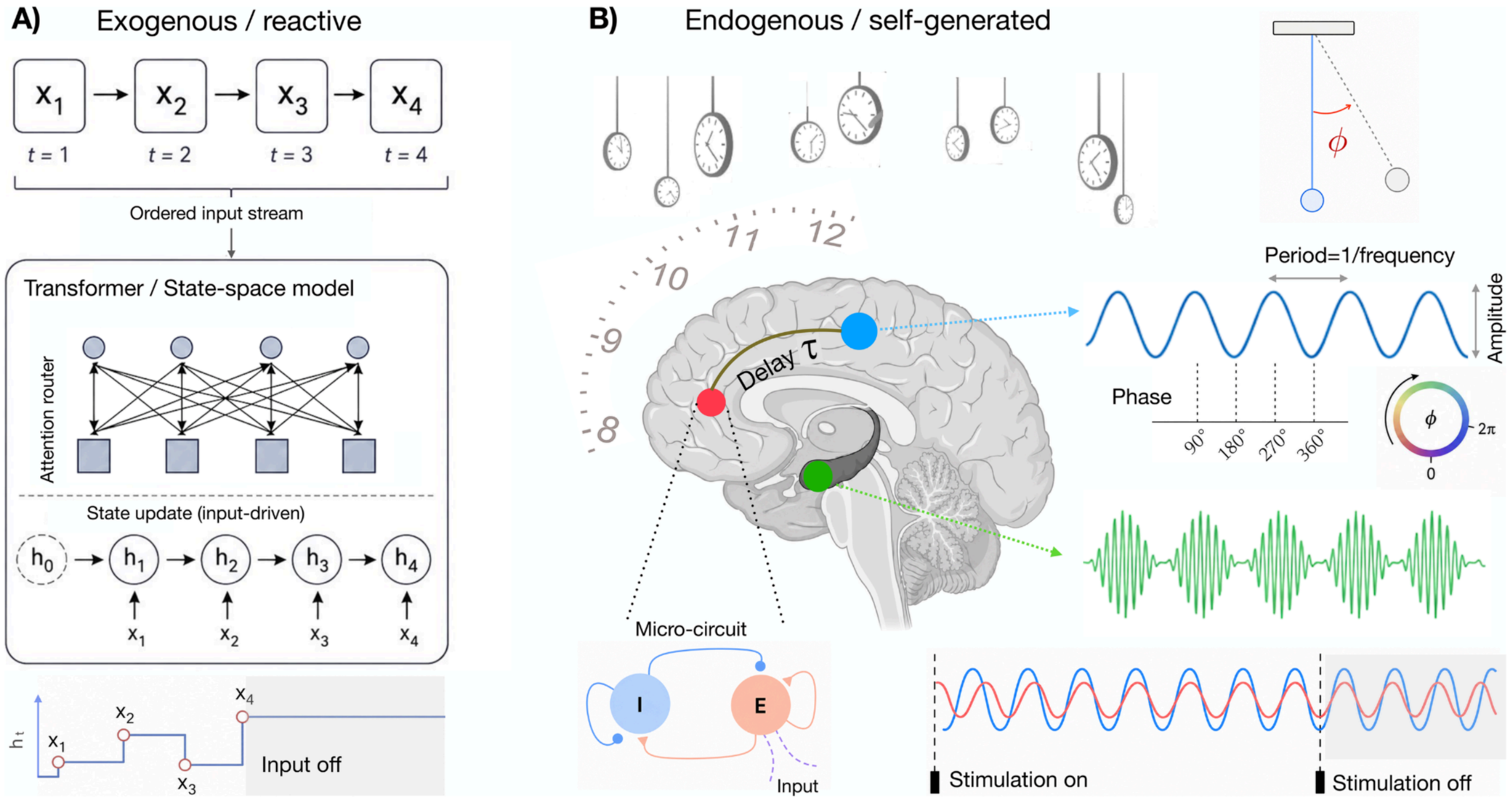

**Figure 1. Exogenous and endogenous temporal computation**. **(A)** In standard sequence-processing architectures such as transformers or discrete-time state-space models, computation is driven by the ordered input stream. Attention weights or internal states denoted by $h_t$ at time $t$ are computed from the current and previous inputs $x_i$. Once the input stream stops, these architectures do not autonomously advance their computational states in real time. **(B)** A cortical circuit generates rhythms endogenously: recurrent excitatory–inhibitory (*E*–*I*) connections can sustain oscillatory activity even in the absence of structured input. A conduction delay $\tau$ between two regions gives rise to a phase offset $\phi$ between their oscillatory activities. External input can reset or entrain the ongoing rhythm (e.g., causing a phase shift), but when it ends, the system keeps cycling.

## Brains fail the timing-invariance test

We defined a reactive system by timing invariance: it computes the same output from the same input, whenever that input arrives. Brains fail this test in behavior [30]. Identical near-threshold targets are detected more often at one phase of the ongoing alpha rhythm than at the opposite phase [31, 32]. Covert attention alternates between locations at 4–8 Hz [33, 34], with matching theta activity in the frontoparietal attention network even during nominally sustained attention [35], and reaction times during recall from sequential working memory fluctuate at theta frequency with encoding-to-recall delay [36]. Behavioral rhythms become visible in trial averages when an event such as a cue, a sound or a movement aligns phase across trials [37]. In this account, the event resets an ongoing rhythm rather than creating it, which is the relation Box 1 requires. However, the evidence is not uniform. A preregistered study did not reproduce the alpha-phase effect on detection [38], the spectral analyses used to detect behavioral rhythms can mistake aperiodic temporal structure for oscillation [39], and the link between individual alpha frequency and the temporal resolution of perception [40] did not hold in a rigorous multi-day test [41].

Where the effects hold, these findings establish what the timing-invariance test asks and no more: the internal temporal process is computationally operative, because its evolving state changes how identical input is processed. This violates timing invariance, but does not establish an advantage. Taken at face value, phase-dependent detection is a cost, since targets are more likely to be missed at the unfavorable phase. Yet one account does assign rhythmic sampling a function: theta-rhythmic attention alternates between sampling the attended location and readiness to shift, trading stability against flexibility through temporal alternation rather than parallel resource allocation [42]. Whether internally timed solutions of this kind outperform reactive ones is the question the rest of this article takes up.

## Four computational functions of endogenous temporal structure

The brain organizes computation in time through oscillations (rhythmic fluctuations in neural activity) and structured delays (lags in signal transmission across neural pathways). Our question is functional, not biomimetic: not whether AI architectures should resemble brains, but whether they need similar temporal operations. It remains an open question whether these are incidental to biology rather than a general computational principle that artificial systems need to adopt explicitly. Sequence-processing models represent temporal structure in response to input: *attention mechanisms* perform content-based routing (selecting pathways as a function of inputs) [43]; *recurrent neural networks* maintain temporal state (carrying summary information from previous inputs) [44]; *positional encodings* specify absolute or relative sequence position (tracking order through a fixed function of position) [3]. In each case, temporal structure is derived from the input rather than generated internally to segment, reset, or modulate information processing. Could input-driven architectures be sufficient to replicate the computational functions of endogenous temporal structure? We argue that they may fall short in four specific ways, each discussed in a section below. The putative gaps and a prediction that would test each case are summarized in Table 1.

### *Phase-dependent routing*

In mainstream AI architectures, routing mechanisms are typically input-dependent: they change with every input, but remain reactive; no autonomous internal process drives them. In transformers, routing between representations—i.e., controlling how hidden vectors encoding input features influence each other—is computed from each input using learned projection matrices fixed after training. The routing pattern (attention weights) is recomputed as inputs are processed. The original transformer uses sinusoidal functions [2] to represent sequence order without additional learned parameters, and many modern variants use position-dependent rotations to encode relative distances within a sequence [4]. However, these architectures encode static positioning rather than autonomously gating which representations communicate. *Mixture-of-experts* models add sparse activation, processing each input through only a selected subset of sub-networks called experts to avoid engaging all parameters. Their selection depends on the current input representation; the router has no recurrent state of its own [45]. State-space models maintain a recurrent internal state updated by each new input and learned state transitions; even in selective variants where the update is content-dependent, the state is updated as the input advances [46]. In each case, computation is driven by what the network receives; none of these explicitly implements an internal temporal process to selectively gate information. Neural oscillatory dynamics implement a different kind of routing. They persist even in the absence of input, and the current position within each cycle, referred to as the phase, dynamically modulates interactions across the network [47].

Under the ***communication through coherence (CTC)*** hypothesis [11, 20], selective routing of information among neural networks is achieved by phase-dependent synchronization (including nonzero phase lags) of their oscillatory activity. A specific phase relationship between two regions dynamically renders their anatomical connections functionally effective, whereas a shift to an unfavorable phase relationship renders them ineffective (see Fig. 2). This phase relationship is not imposed by the input, but rather generated endogenously, and can shift on a millisecond timescale, without changes in the underlying anatomical connectivity. By opening brief excitability windows at preferred phases, the same neural circuit can realize different patterns of effective connectivity [48]. This routing is both flexible and self-generated [49], a combination no mainstream AI architecture explicitly implements.

Empirically, gamma-band synchronization between cortical regions modulates the efficacy of neuronal interactions [50]. Moreover, feedforward and feedback signaling are carried by different frequency bands (bottom-up: theta–gamma; top-down: alpha–beta), indicating that hierarchical modulation is associated with distinct rhythms [51]. Neural rhythms carry a further functional role: fast rhythms (>30 Hz) determine “which” populations communicate, and slow rhythms (<25 Hz) select “when” they are receptive [52]. This operates in two complementary ways. Alpha oscillations suppress excitability in task-irrelevant regions through power modulation [53], whereas delta and theta rhythms open windows of high excitability to behaviorally relevant events through phase alignment [54]. These two mechanisms—coherence-based routing and amplitude-based gating—have been proposed to operate jointly in a unified framework based on nested oscillations [52]. Across fast and slow rhythms, then, the same principle holds: the control of communication is dynamic, set by the ongoing rhythm rather than only by anatomy. Whether coherence reflects a causal routing mechanism or a downstream signature of communication remains uncertain. Either way, whether mechanisms such as attention and mixture-of-experts can approximate the function of phase-dependent routing remains to be tested empirically.

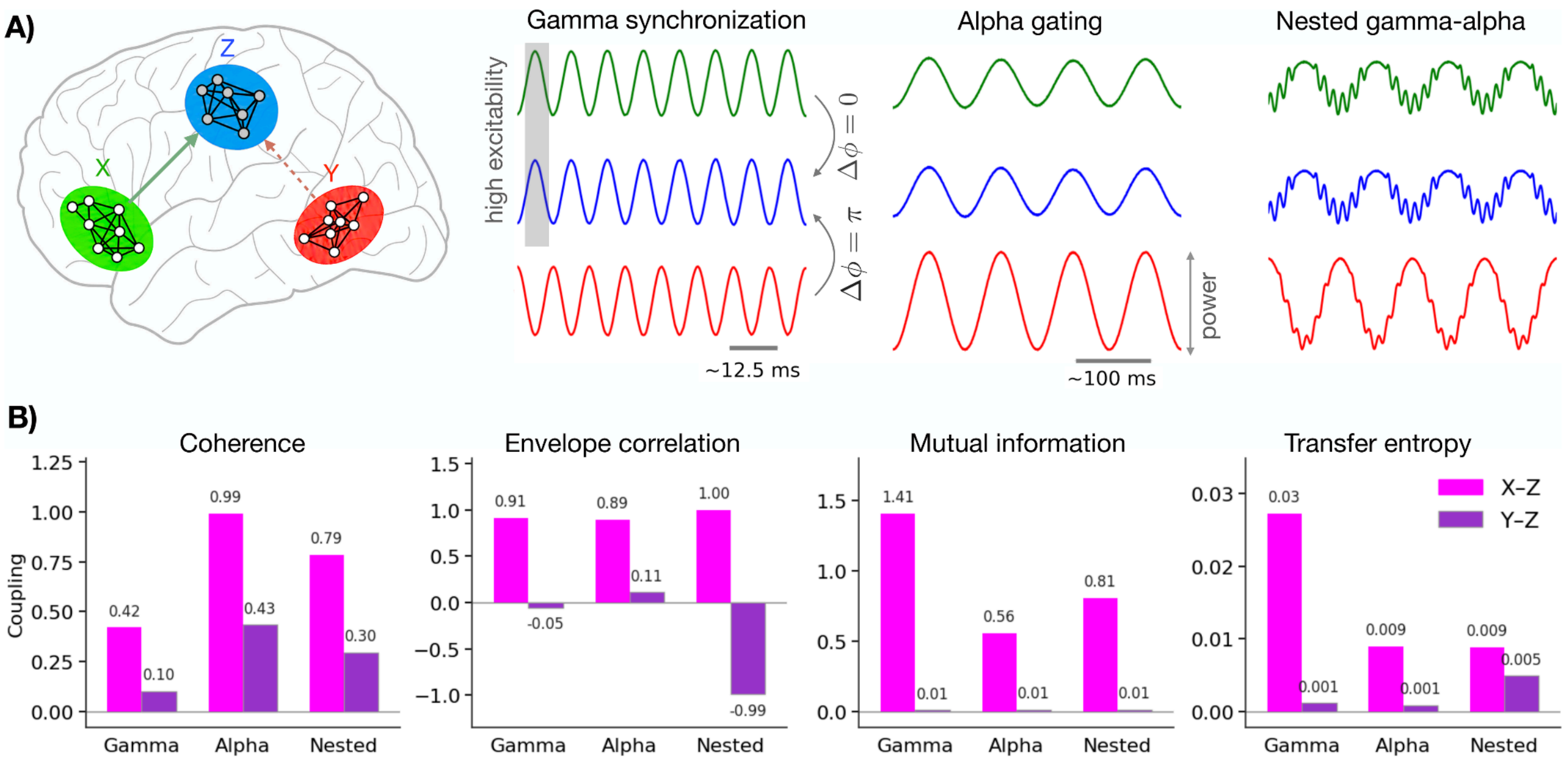

**Figure 2. Coherence-based routing, amplitude-based gating, and their integration through nested oscillations. (A)** Left: schematic illustration of three interacting cortical regions: sender *X* (green), receiver *Z* (blue), and competitor *Y* (red). Right: Gamma synchronization: The pair *X* and *Z* share high-excitability windows (in gray) via phase alignment (*Δφ = 0)*, whereas *Y* is antiphase with *Z* (*Δφ = π*). Alpha gating: differences in power across populations modulate their excitability. Nested oscillations: gamma amplitude is modulated by alpha phase, integrating phase-selection and amplitude-gating. **(B)** Measured coupling between *X–Z* (magenta) and *Y–Z* (purple): coherence captures phase alignment; envelope correlation the co-modulation of amplitude; mutual information the shared variability; and transfer entropy the directed statistical dependency. Each of the metrics shows stronger coupling over *X–Z* than *Y–Z.* For details see *Supplementary section 1*.

### *Delay-structured representation*

A second, often overlooked dimension concerns *transmission delays.* Axonal conduction delays are not uniform: they range from under a millisecond in local pathways [55], through a few milliseconds in thalamo-cortical projections [56], to tens of milliseconds in long-range cortico-cortical pathways [57]. This heterogeneity interacts directly with oscillatory dynamics: a specific time delay at a given frequency produces a specific phase offset between sender and receiver, which under CTC modulates communication efficacy [48]. Beyond such pairwise coupling, delayed neural signaling can shape neural activity across scales, from single-neuron firing [58], through stimulus responses in neural circuits [59], to collective network dynamics arising from short- and long-range interactions [60, 61, 62, 63]. For instance, delays modulate phase relationships between regions—with antiphase locking occurring at higher frequencies and across long-range connections—thereby driving synchronization [23]. Delays can support zero-lag synchronization through dynamical relaying [64] and shape communication through synchronization [65]. In networks of coupled oscillators (such as the Kuramoto model), delays determine stability and the repertoire of collective dynamics, including phase-locked states, traveling waves, and chaotic behavior [23, 66]. In plastic neuronal networks, delays can additionally influence synaptic connectivity patterns [67, 68]. Crucially, these effects arise from delayed coupling itself and do not depend on any particular account of how coherence regulates communication.

Delays reshape computation, not just communication. With time delays and spike-timing-dependent plasticity, a recurrent network can self-organize into ***polychronous neuronal groups***—precise, time-locked, but asynchronous firing patterns that encode discrete representations [69]. The plasticity takes an initially unstructured network and selects firing patterns. Due to heterogeneous delays, spikes emitted at different times can arrive at a common postsynaptic neuron within a few milliseconds, initiating a cascade of further spikes. Each recruited neuron sends delay-adjusted spikes to downstream targets, and recurrent connectivity propagates the resulting pattern. Through this interplay between the delays and plasticity, the same neuron can participate in many such patterns with different timings, so the number of coexisting groups can far exceed the network size [69]. This combinatorial capacity provides a rich substrate for memory storage, pattern recognition, and cognitive processes. By contrast, sequence-processing architectures such as transformers abstract away time delays. State-space models discretize a learned continuous-time system, whose finite-dimensional state can approximate delays. Whether positional encodings, recurrent memory, temporal filtering, or continuous-time models can reproduce the computational functions of connection-specific delays remains an empirical question. Box 2 develops this argument and proposes a direct test.

### Box 2. Delays as combinatorial resource

Consider a network of excitatory–inhibitory (*E–I*) modules in which each connection $i \to j$ carries its own transmission delay $\tau_{ij}$. Formally, this is a system of *delay-differential equations*, whose state space is infinite-dimensional even for a finite number of modules [70]. At an operating frequency $f$, each delay imposes a phase lag $\phi_{ij} = 2\pi f \tau_{ij} \pmod{2\pi}$ between sender and receiver. According to the ***communication through coherence (CTC)*** hypothesis, this delay-induced offset modulates communication efficiency: signals arriving during phases of high excitability are transmitted more effectively than those arriving outside these windows. The width of these windows depends on *E–I* dynamics: faster recruitment of inhibition by excitation narrows the phase interval during which inputs can drive postsynaptic activity, making conduction delays more selective in gating postsynaptic responses. Thus, the distribution of delays, not only synaptic weights, shapes effective communication via frequency and phase relationships [59, 71]. While the CTC emphasizes phase synchronization between neuronal populations within a frequency band, cross-frequency coupling involves an intricate dialogue between different brain rhythms across spatial and temporal scales. Independent of either framework, transmission delays also shape collective network dynamics, including synchronization patterns [72, 73].

In classical recurrent networks, asymmetric connectivity and nonlinear feedback can give rise to fixed-point and limit-cycle attractors, spatial sequences, and associative memory, but storage capacity generally scales linearly with network size [74, 75]. In coupled-oscillator networks the limit is even sharper: homogeneous delays can drive the dynamics toward global synchronization. Heterogeneous, connection-specific delays can break this symmetry, rendering the system multistable [76]. They set the phase-lag structure between coupled regions: the coexistence of stable attractors, each with its own basin of attraction and synchronization pattern. This yields a substantially expanded dynamical repertoire, extending computational and memory capacity beyond the limits imposed by network size.

With spike-timing-dependent plasticity, delay-coupled networks self-organize into ***polychronous neuronal groups (PNGs)***, which are spatiotemporal firing sequences within groups of neurons. During training, activity in such networks can evolve from delta oscillations through irregular firing to gamma oscillations. Despite distinct emission times, spikes converge on a common postsynaptic target because delays compensate for the timing offsets [69]. Reversing their temporal order can instead direct coincident arrivals to another target, illustrating how delays enable target-specific routing. Even without structured input, background noise can trigger a PNG; however, due to continuous plasticity it can suddenly disappear. PNGs share the following characteristics: ***Transience:*** each activation is not a fixed-point attractor but an ordered sequence of firing, typically lasting up to 100 ms. ***Context dependency:*** Different inputs recruit groups with different spatiotemporal profiles over different sets of neurons. ***Reproducibility:*** triggering the anchor neurons under different perturbations regenerates the same sequence with sub-millisecond precision (see Fig. 3). Crucially, the number of coexisting groups can far exceed the neuron count, yielding a combinatorial capacity unavailable to instantaneous communication. These counts, however, depend strongly on the minimum chain length chosen to define a group, and enumeration cost grows rapidly with network size. Whether heterogeneous delays are responsible for this capacity is directly testable: Hold the neuron count, weights, connectivity, inputs, and neuronal parameters fixed, and compare the intact delay against instantaneous and shuffled-delay controls. If capacity declines under delay shuffling or collapses under removal, it is attributable to the delays; if not, the claim is falsified (see Fig. 4).

In artificial neural networks, connection-specific transmission delays can be implemented using differentiable ring buffers with linear interpolation between adjacent positions, allowing gradients to propagate through continuously parameterized delays. Such operators incur modest computational cost when delays are sparse; dense connection-specific delays scale less favorably in memory and computation overhead, whereas state-space alternatives approximate delayed signal propagation through learned recurrent dynamics rather than representing each delay explicitly. Learnable transmission delays have also been explored in spiking neural networks [77]. The engineering challenge is therefore not the availability of computational primitives, but their efficient implementation while preserving the temporal relationships relevant to computation.

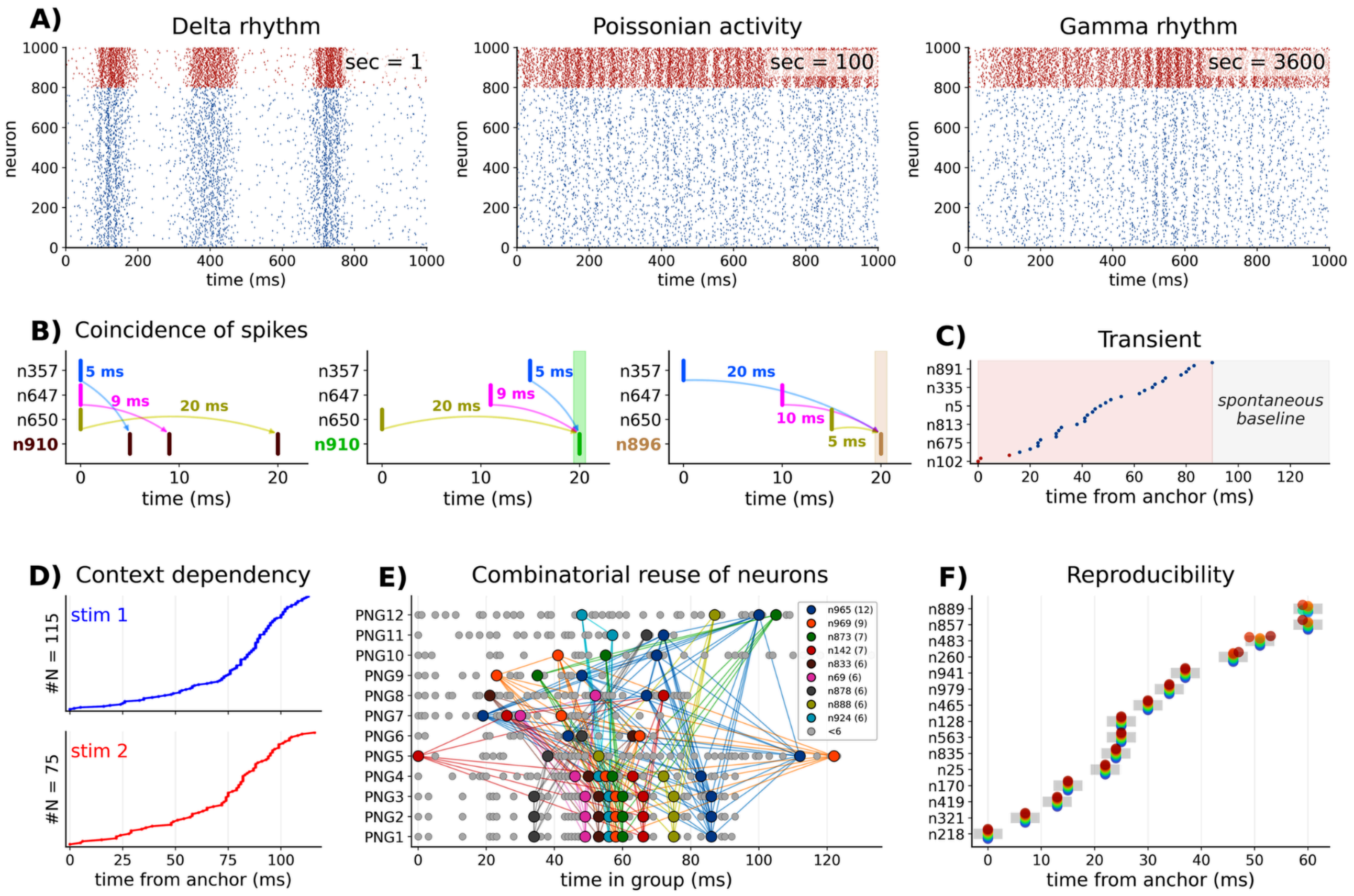

**Figure 3. Polychronous neuronal groups.** A recurrent network of spiking neurons ($N$=1000) with spike-timing-dependent plasticity was trained for 18000 sec. Excitatory delays were uniformly distributed over 1–20 ms, while inhibitory delays were fixed at 1 ms. Excitatory and inhibitory neurons (4:1 ratio) followed regular-spiking and fast-spiking dynamics, respectively. **(A)** The spiking activity of excitatory (blue) and inhibitory (red) neurons at different training stages: initial delta oscillations (sec=1), relatively uncorrelated Poissonian activity (sec=100), and emerging gamma oscillations (sec=3600). **(B)** Delay compensation underlying polychronization: presynaptic spikes arrive at different times and fail to reach the firing threshold (left), whereas heterogeneous delays compensate for emission-time offsets so that spikes converge on a common target (middle). Reversing the emission order produces coincidence at another neuron (right). This demonstrates how delays transform temporal order into target-specific routing. **(C)** Transient activation: sequential cascades lasting up to 100 ms, after which activity drops to the spontaneous level. **(D)** Context dependency: two different stimuli yield different spatiotemporal profiles over different sets of neurons. **(E)** Combinatorial reuse of neurons: in 12 groups (one per row), 476 participations are realized by 262 unique neurons at different relative times, linked as a chain of cascades (arrows). **(F)** Reproducibility: triggering the anchor neurons of one group over 10 trials, each with an independent realization of background noise, reactivates the same sequence with sub-millisecond precision (mean deviation time $|\Delta| \approx 0.06$ ms relative to the noise-free reference). For details, see *Supplementary section 2*.

To test whether time delay is directly responsible for combinatorial capacity, we performed a delay-ablation test in recurrent and spatial spiking networks (Fig. 4A,B). Both networks, with regular- and fast-spiking dynamics for excitatory and inhibitory neurons, used the same plasticity rule (Fig. 4C), but differed in how their delays were assigned: uniformly in the recurrent network and according to connection distance in the spatial network (Fig. 4D,E). We distinguish capacity from memory retrieval. Capacity is a structural quantity: the number of distinct groups a network can support given its connectivity, weights, and delays. Memory retrieval is functional: the fraction of sequences that are successfully reactivated at their original timing after perturbation. This distinction becomes clear after delay shuffling. Capacity can be partially preserved because strong synaptic connections via plasticity, combined with a delay matrix, can generate new coincidence-driven cascades. In contrast, memory retrieval depends on the specific temporal relationships underlying learned sequences, providing a more direct signature of the delay structure. This is also reflected in an energy-like landscape of pattern similarity (Fig. 4F), where larger groups form deeper local minima under the original delays. Shuffling flattens the landscape, indicating disrupted structured replay without silencing ongoing network activity.

The group counts fall rapidly as chain length increases (Fig. 4G, left), because longer chains of cascades require more delay-matching constraints. The number of candidate triplets generating coincident arrivals grows rapidly with network size (Fig. 4G, right), making the enumeration costly in large networks. The ablation results in both tested networks show that delay shuffling degrades retrieval far more than capacity (Fig. 4H), particularly for longer chains ($L \geq 7$), where retrieval is completely abolished while some capacity remains. The capacity drops more sharply in the spatial network, where shuffling disrupts the relationship between connection distance and delay, despite preserving the overall delay distribution.

Instantaneous spike transmission removes the delay-related temporal dispersion that supports the asynchronous patterns underlying PNGs. Although a small fraction of shorter chains ($L \geq 3$) still meet the retrieval threshold, longer chains fail in both networks. Capacity collapses to zero under removing delay, in both networks and regardless of chain length. These simulations provide a proof-of-concept that connection-specific delays can encode information beyond synaptic weights and connectivity.

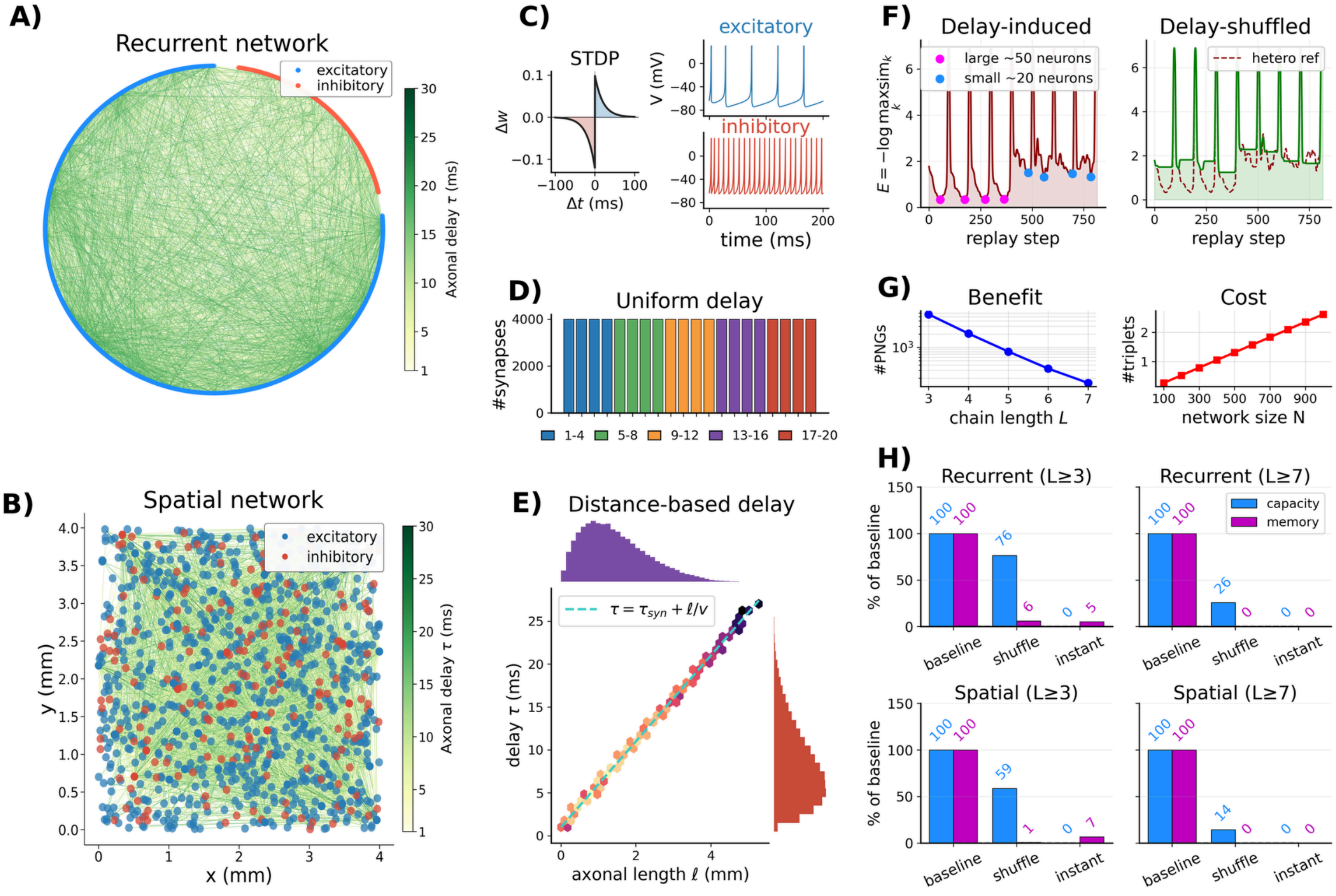

**Figure 4. Delay-ablation test.** Two networks are trained with the same plasticity and neural dynamics but different delay structures. **(A)** Recurrent spiking network with heterogeneous delays. **(B)** Spatial network of spiking neurons distributed over a 4×4 mm sheet and coupled via distance-dependent delays. **(C)** A pair-based spike-timing-dependent plasticity rule applied only to excitatory weights. Excitatory and inhibitory neurons (4:1 ratio) followed regular-spiking and fast-spiking dynamics, respectively. **(D)** Delays in the recurrent network are heterogeneous and uniformly distributed (1–20 ms). **(E)** Delays in the spatial network are inhomogeneous ($\tau_{ij} = \tau_{syn} + \ell_{ij}/v$, with $\tau_{syn}$ = 1 ms and $v$ = 0.20 mm/ms), with a right-skewed distribution (mode ≈ 5 ms, with a tail to ~30 ms). **(F)** An energy-like representation is calculated as the negative log of the maximum cosine similarity to the reference patterns. The original-delay landscape shows minima associated with smaller (~20 neurons; left, blue) and larger (~50 neurons; left, magenta) groups. Delay shuffling (right, green) lifts these minima relative to the original-delay reference (right, dashed red), particularly for larger groups, indicating disrupted replay of learned patterns. **(G)** The group counts fall rapidly with chain length $L$ (left), whereas candidate triplets grow rapidly with network size $N$ (right). **(H)** Delay ablation separates representational capacity (blue) from memory retrieval (magenta). Bars are shown for recurrent (top) and spatial (bottom) networks at chain-length thresholds $L \geq 3$ (left) and $L \geq 7$ (right). Delay shuffling partially preserves capacity but strongly impairs retrieval for shorter chains and abolishes it for longer chains in both networks. Removing delays collapses capacity to zero at both chain-length thresholds. Moreover, a small fraction of shorter sequences can still be retrieved whereas longer chains fail in both networks. For details, see Supplementary section 2.

### *Cross-frequency multiplexing*

A third consideration concerns how multiple items can be maintained simultaneously in working memory. This involves holding their representations in a readily accessible form and actively transforming them over short periods, despite limited capacity and changing environmental demands. The underlying mechanisms must therefore explain not only how items are represented, but also how their identities are separated. Neuronal oscillations provide an internal timing reference for this organization. Both low- and high-frequency oscillations coexist in many brain regions and interact with each other, a process called *cross-frequency coupling* [78, 79] that has been implicated in functions such as multi-item representation, long-range communication, and multiscale integration of information [17].

The *theta–gamma neural code* [15] posits that the brain organizes information—such as items in working memory or spatial sequences—by nesting fast gamma oscillations within slower theta cycles, providing a form of time-division multiplexing [80]. In this scheme, a subset of cells that co-fire during a gamma cycle forms a cell assembly whose spatial pattern of activity represents a given item. The assemblies active in different gamma cycles encode distinct items at successive phases of the underlying theta rhythm. Thus, the position of each gamma cycle (or equivalently, its theta phase) specifies the serial order of the items [81, 82, 83]. Empirical studies link theta–gamma coupling to working-memory load and performance in the hippocampus [84], impaired coupling in schizophrenia [85], and age-related deficits in associative memory [86]. That such oscillatory dynamics can implement multiplexing is shown directly: adding dynamical refraction with oscillatory inhibition to a trained network enabled it to segregate competing inputs into a sequence, a task it could not otherwise solve [87].

Standard transformer architectures have no built-in analog of such hierarchical temporal organization. Sequence order is instead supplied through positional representations, while interactions among sequence elements are mediated by attention [2, 3]. Although this flat, parallel representation can be extended through positional interpolation and sliding windows, the architecture inherently lacks nested structures for temporal integration, especially when scaling to long-range context. State-space models [21, 46] come closer by using input-dependent state transitions and selectively forgetting, but multi-scale conflict remains a challenge even for their selective variants. This is precisely the function that cross-frequency coupling implements. Whether multi-frequency oscillations offer computational advantages for sequence-processing models remains a practical question.

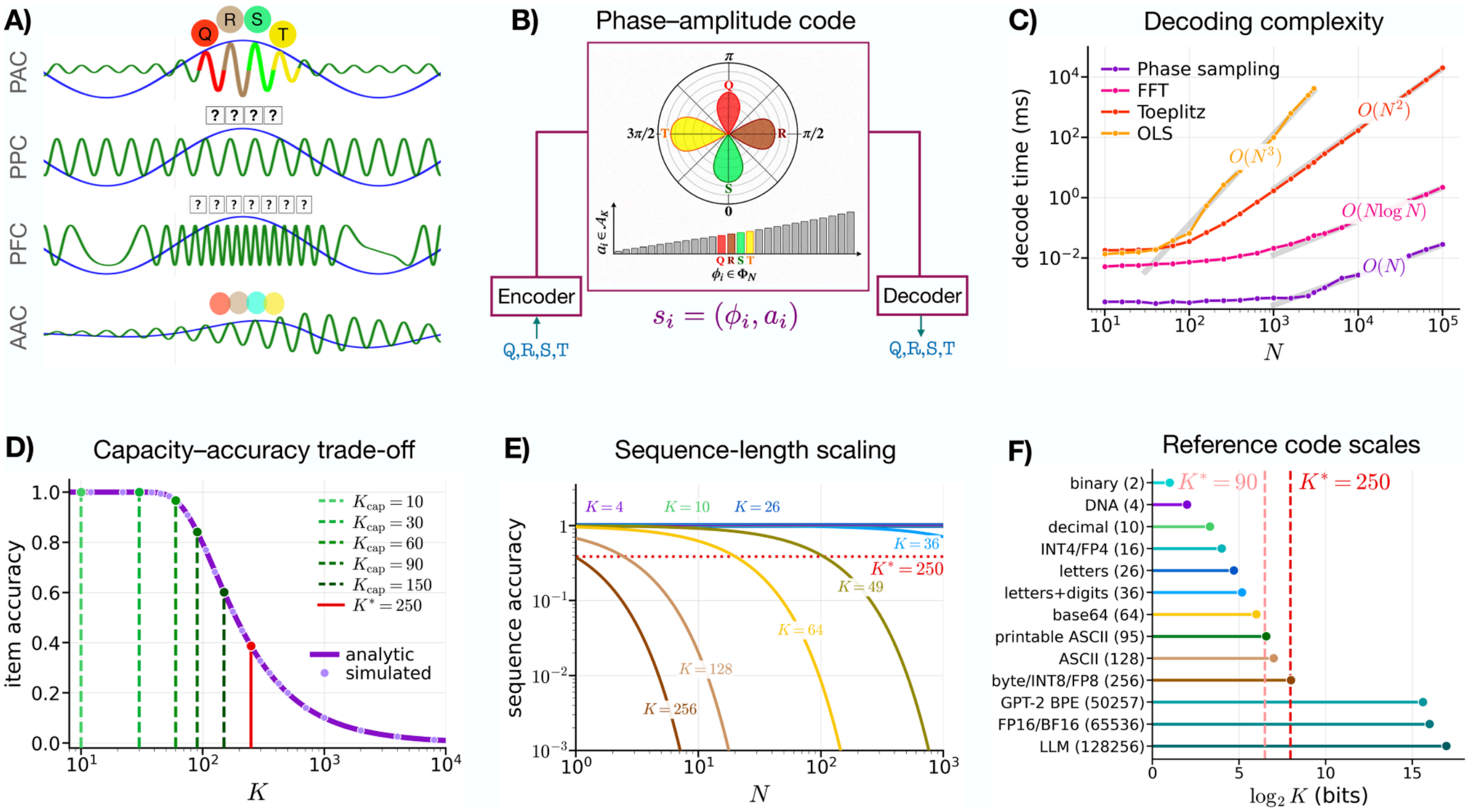

**Figure 5. Theta–gamma sequence coding. (A)** Canonical cross-frequency coupling motifs. Phase–amplitude coupling (PAC) organizes sequential representations by nesting gamma oscillations within a theta cycle, whereas other motifs lack amplitude gating or stable temporal slots. **(B)** An abstract phase–amplitude code, $s_i = (\phi_i, a_i)$, with phase specifying ordinal position and amplitude encoding item identity. **(C)** Decoding complexity as a function of sequence length $N$, showing the favorable scaling of direct phase sampling relative to other approaches. **(D)** Item accuracy decreases with dictionary size $K$, as the spacing between amplitude levels decreases. Dashed lines: operational capacities $K_{\text{cap}}$, as the largest dictionary sizes meeting specified accuracy thresholds. Solid line: characteristic noise scale $K^*$ defined as $K^*=R/\sigma_a$, shown for the alphabet code with fixed range $R$=25 and $\sigma_a$=0.1 as the amplitude noise level. **(E)** Exact-sequence accuracy decreases with $N$, with faster degradation for larger $K$, as item errors accumulate across positions. Dotted line: the per-item accuracy $P_{\text{id}}(K^*=250)\simeq0.4$, shown for comparison. **(F)** Reference vocabularies on a $\log_2 K$ scale, shown against the characteristic noise scale $K^*$. Dashed lines: for digit and alphabet codes $R$=9 and $R$=25, giving $K^*$=90 and $K^*$=250, respectively. For details, see Supplementary section 3.

Fig. 5 illustrates a phase–amplitude coding scheme, inspired by the theta–gamma neural code, in terms of sequence representation, accuracy, complexity, and the limits imposed by finite resolution and noise. Different coupling motifs characterize statistical dependencies between oscillatory phase, amplitude, and frequency, providing a versatile framework for neural coding and information processing (Fig. 5A). Phase–amplitude coupling (PAC), in which the amplitude of faster activity is modulated by the phase of a slower rhythm, provides a temporal framework for organizing sequential representations. This offers the most direct basis for our proposed coding scheme, with phase specifying item position (as neuronal firing can occur at distinct theta phases) and amplitude (presumably reflecting the number of active cells) encoding item identity. Phase–phase coupling (PPC) establishes consistent phase-locking relationships between oscillations, but lacks amplitude gating. Phase–frequency coupling (PFC), in which the phase of a slow rhythm modulates the instantaneous frequency of a faster rhythm, often exhibits amplitude artifacts under the Hilbert transform, and disrupts stability of temporal slots. Finally, amplitude–amplitude coupling (AAC) lacks an explicit phase coordinate for gating and ordering.

Our proposed scheme abstracts the theta–gamma principle into a phase–amplitude coding (Fig. 5B): each item is represented as $s_i = (\phi_i, a_i)$ where phase $\phi_i \in \Phi_N$ specifies ordinal position and amplitude $a_i \in A_K$ encodes identity. Here, $\Phi_N$ denotes the set of $N$ phase slots, while $A_K$ denotes the dictionary of $K$ amplitude levels. The two dimensions impose orthogonal resolution constraints: increasing sequence length $N$ reduces spacing between temporal slots within a fixed cycle, whereas increasing dictionary size $K$ reduces the spacing between amplitude levels within a fixed range. Consequently, repeated identities do not reduce decoding accuracy because each position is evaluated independently. This structure also simplifies decoding (Fig. 5C): the encoded amplitudes can be recovered directly by sampling the signal at corresponding phases, without matrix inversion or iterative reconstruction, yielding approximately linear computational scaling, while FFT-, Toeplitz-, and OLS-based reconstruction incur higher computational costs in this implementation.

The principal limitation arises from finite resolution under noise. The amplitude dimension sets the item-level resolution limit (Fig. 5D). For $K$ uniformly spaced levels over a fixed range $R$, the adjacent levels are separated by $d=R/(K-1)$. As $K$ increases, the overlap between identities increases as their spacing decreases relative to the noise scale, thus increasing the probability of decoding errors. The characteristic scale is defined by $K^*=R/\sigma_a$, where $\sigma_a$ denotes the standard deviation of noise to amplitudes, marks the regime in which the noise scale becomes comparable with adjacent-level spacing. Note that $K^*$ is not a hard capacity bound. Instead, the operational capacity $K_{\text{cap}}$ determines the largest dictionary size that achieves a threshold in item-level accuracy. These item-level errors become more consequential for complete-sequence recovery (Fig. 5E). Exact recovery requires every item to be decoded correctly, so sequence accuracy decreases with $N$, with increasingly rapid degradation for larger $K$. Specifically, $K$ controls item-decoding errors, whereas $N$ determines how these errors accumulate across the sequence. Note that our implementation treats phase and amplitude as independent dimensions and therefore does not explicitly model theta–gamma code, in which theta phase modulates gamma activity. The computational contribution of such phase-dependent gating remains to be investigated.

Finally, we show the characteristic dictionary scales relative to standard symbol sets and language-model vocabularies on a $\log_2 K$ scale, indicating the theoretical number of bits required to identify a symbol (Fig. 5F). The estimated characteristic scales, $K^* \approx 90$ and $K^* \approx 250$, correspond to approximately 6.5 and 8 nominal bits per amplitude-coded item, respectively, spanning typical character-level symbol sets and approaching the 256-symbol cardinality of byte-level coding. Note that these values characterize the noise-limited resolution of the scalar amplitude code, rather than the maximum dictionary size that can be achieved at a specified decoding accuracy. Larger token vocabularies require substantially more distinct identities than a single noisy amplitude can accommodate at these scales. This is not a direct comparison of model capacity: modern transformer-based models represent token identity using high-dimensional learned embeddings rather than a single scalar amplitude, while sequence order is specified through positional encodings. Moving from a scalar to a distributed representation allows symbol identities to be encoded as high-dimensional vectors, increasing the available representational space without requiring progressively finer amplitude resolution in a single dimension. The comparison therefore illustrates the trade-off between scalar resolution and representational dimensionality, without implying equivalent memory capacity or computational power.

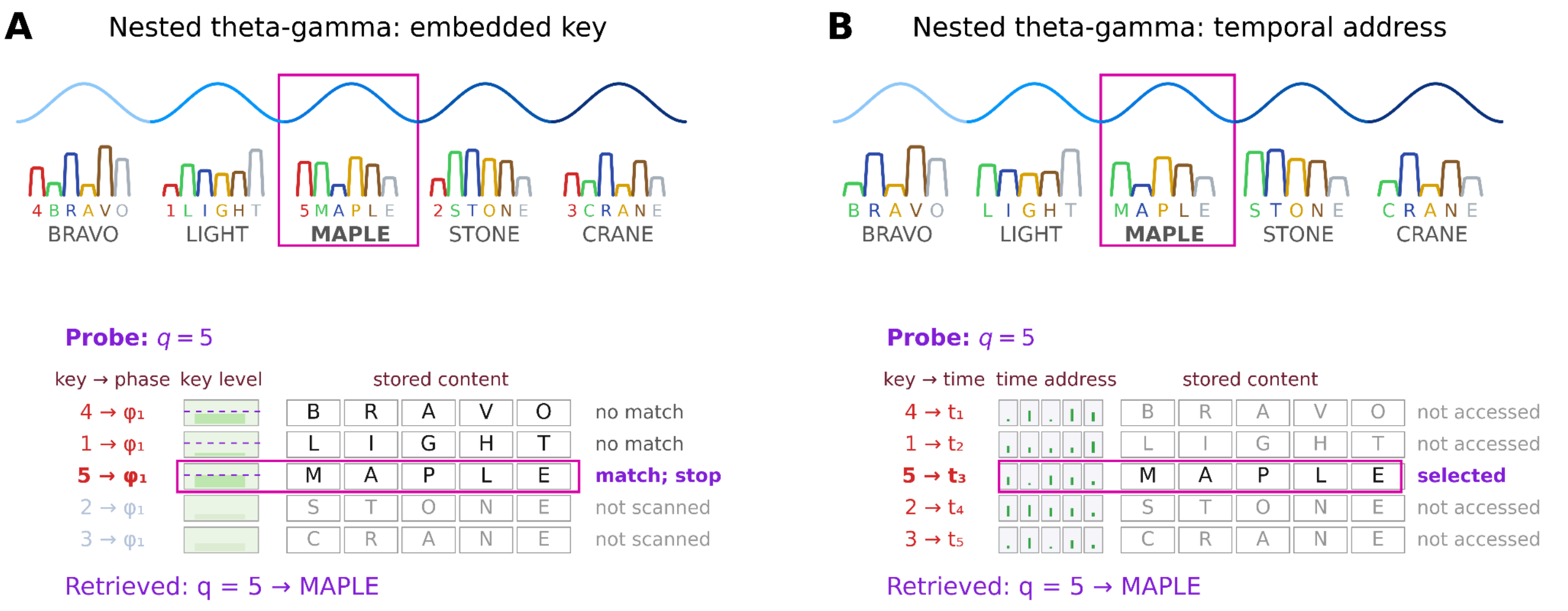

**Figure 6. Embedded-key retrieval versus temporal addressing using nested oscillations. (A)** Embedded-key retrieval: Each theta cycle holds an amplitude-coded key (digit) in the first nested gamma subcycle, followed by content items (letters). During retrieval, a probe (here $q$ = 5) is compared sequentially with stored key levels using a predefined matching tolerance. The first two keys fall outside the tolerance and are rejected; the third matches and its associated sequence (*MAPLE*; magenta outline) is retrieved, while the remaining sequences are not examined. **(B)** Temporal addressing: Each theta cycle holds content only, without an embedded key in the nested gamma subcycles. Instead, an external key-to-time mapping converts the probe into the temporal address of the corresponding sequence. Retrieval goes directly to that sequence (*MAPLE;* magenta outline), without scanning others, assuming the mapping and direct temporal access are available.

Beyond encoding item identity and serial order, temporal structure can also support selective memory retrieval. Extending the theta–gamma code, nested oscillations can serve as an address for retrieving sequences (Fig. 6). In *embedded-key retrieval*, each theta cycle contains an amplitude-coded key placed in the first nested gamma subcycle (i.e., early theta phase), followed by a sequence of items (Fig. 6A). A probe initiates a self-terminating serial search that stops at the first key satisfying a matching criterion. This avoids examining subsequent sequences, but retrieval cost depends on the target's position, and similar or noisy keys can result in false matches or failed retrieval. In *temporal addressing*, the key is not embedded in the stored sequence; instead, an external key-to-time mapping converts the probe into a temporal address (e.g., a theta-cycle index), allowing the corresponding cycle to be selected directly without scanning other sequences (Fig. 6B). This could reduce the number of memory accesses but requires reliable mapping and direct temporal access, while remaining vulnerable to timing errors or clock drift. Thus, any advantage of direct temporal retrieval depends on a reliable mapping, with its storage and lookup costs accounted for. The two mechanisms need not be mutually exclusive. Their combination suggests a hierarchical form of multiplexing, in which the temporal location of a slower theta cycle identifies a stored sequence, while faster gamma subcycles provide access to its individual items. Whether such organization improves retrieval efficiency and whether phase can serve as a computational variable for transforming sequence representations remain to be tested.

***Oscillation-gated offline consolidation***

A fourth consideration concerns the periods between processing episodes, when the system is not engaged with input. Biological memory is reorganized during these offline states, and the reorganization is timed by endogenous oscillatory events [88]. Hippocampal sharp-wave ripples (SWRs) are brief population events, typically 40–100 ms long, carrying a fast oscillation at 110–200 Hz in rodents [89] and around 80–100 Hz in human intracranial recordings [90]. They occur during non-REM sleep and quiet wakefulness, carry compressed replay of recent experience [89], and their selective disruption during sleep impairs subsequent memory [91]. During awake rest they can also express sequences along future or not-yet-traveled paths [92, 93]. During sleep, ripples sit within a hierarchy: spindles are grouped by the up-state of cortical slow oscillations and in turn cluster ripples in their troughs [90], a nesting proposed to provide temporal windows for hippocampal-to-cortical transfer.

Two causal experiments show that this timing matters, not only the occurrence of replay. Reinforcing the coupling of ripples with cortical delta waves and spindles by closed-loop stimulation after learning produced next-day recall in rats that otherwise performed at chance, and the benefit disappeared when the stimulation was delivered at a variable delay [94]. Prolonging ripples that the network initiated itself improved memory, whereas ripples induced at random times did not, and carried little spatial information [95].

Most sequence-processing models have no endogenously timed equivalent. Base models do not adjust their parameters between successive inputs; any update requires an externally scheduled training phase. Retrieval-augmented systems fetch from an auxiliary memory only when an input or an external controller triggers them [96], and continual-learning methods replay stored or generated experience on a schedule set by the training loop [97, 98]. Replay content, however, is not the missing piece. Prioritized experience replay already selects samples by their expected value for learning [99], and the same utility-based prioritization accounts for much of the content of hippocampal replay without invoking oscillations [100]. We therefore regard this as the weakest of the four functions, and the burden on it is specific. Once replay content is matched, what remains for oscillations is the coordination of when a fast learner and a slow learner interact, the problem that *complementary learning systems* were proposed to solve [101]. In the brain this coordination is set by the phase relations among slow oscillations, spindles and ripples, and the experiments above show that disrupting it removes the benefit. The corresponding test in an artificial system pairs a fast episodic store with a slow parametric learner that receives the same replayed content. In one condition, transfer is gated by coupled internal rhythms; in the other, it follows a fixed or random schedule matched in replay count.

**Table 1.** Four computational functions of endogenous temporal structure, their biological evidence, corresponding gaps in mainstream sequence-processing AI models, and the predicted advantages against a reactive baseline under resource-matched conditions.

| Function | Biological evidence | Gap in AI models | Predicted advantage |
|---|---|---|---|
| **Phase-dependent routing** | Phase relations between rhythms predict inter-areal transfer without changes in anatomy; whether coherence causes or reflects communication is debated | Routing is recomputed from the current input; no internal phase variable controls which modules communicate | Context-dependent routing in which the rule must be held internally rather than read from the current input: higher accuracy at matched state size, with the gap growing with the number of contexts |
| **Delay-structured representation** | Conduction time delays, ranging from sub- to tens of millisecond shape network dynamics; with synaptic plasticity they support polychronous groups that outnumber the neurons | no connection-specific time delays; temporal offsets enter only via position or lags shared across connections | Recognition and recall of patterns defined by inter-event intervals: learned delays outperform the baseline at matched parameters, with the gap growing as the activity propagates deeper through the networks |
| **Cross-frequency multiplexing** | Gamma activity nests within theta cycles; item-selective activity appears at theta/alpha phases matching serial position | Order is supplied by position; no nested internal cycle assigns items to slots | Ordered recall of items with a fixed state budget: higher accuracy than fixed-state baselines, and a capacity limit that tracks the frequency ratio |
| **Oscillation-gated consolidation** | Slow oscillations are hierarchically nested; reinforcing their coordination enhances consolidation, but not when timing is jittered | Replay is scheduled by the training loop; no internal rhythm coordinates fast and slow learners during offline learning | Continual learning with offline periods: higher retention and sample efficiency than prioritized replay scheduled by the training loop, matched in replay content and count |

## The computational hypothesis

The preceding arguments motivate a falsifiable claim to be tested under resource-matched conditions. Its scope follows from Box 1: tasks in which input arrives as discrete events at irregular intervals, where either the intervals carry information or computation must continue between events. Its predicted advantages for the four computational functions of endogenous temporal structure are summarized in Table 1. When input is sampled on a fixed clock, a reactive model can infer elapsed time from sequence position, so no advantage is predicted on this basis. Together, these arguments converge on the following hypothesis:

*(i) A system with endogenous temporal structure will show advantages for each function over a resource-matched reactive system that receives event timestamps as input.*
*(ii) Both systems will perform equivalently on atemporal tasks, within a fixed margin.*
*(iii) These advantages will shrink or vanish when the temporal structure is disrupted.*
*(iv) For the three oscillation-based functions, the system will also outperform the same architecture driven by an equivalent external oscillation.*

Each clause is a separate falsifiable prediction as follows:

(i) Giving the reactive baseline timestamps is essential: without them, failure on timing-dependent tasks could simply reflect missing temporal information, rendering the failure uninformative by construction. Matching memory capacity matters as well, because a transformer's attention cache grows with sequence length while a recurrent state remains fixed-size; without controlling for this asymmetry, ordered-recall comparisons may favor the transformer for reasons unrelated to time.

(ii) The atemporal control tests for non-temporal confounds, such as extra capacity or easier optimization, which would otherwise manifest as uniform gains across all tasks.

(iii) The ablation tests whether the advantage stems from the temporal mechanism as it is diminished by selectively disrupting its operation (see Box 3).

(iv) The last clause tests what endogenous generation adds. If the first three clauses hold and the last fails, the benefit comes from temporal structure rather than from generating it internally. This would weaken the hypothesis while still establishing a computational utility of temporal structure, making either outcome informative.

The claim concerns computational function, not biological implementation. Artificial systems need not use ion channels or action potentials; oscillations and delays are candidate mechanisms whose necessity and optimality are part of what the tests decide. Nor does the hypothesis exclude architectures that already exhibit autonomous temporal dynamics. Finally, the temporal mechanisms considered here operate largely on millisecond-to-second timescales, although their effects, particularly on consolidation, may extend over longer periods. The reported failures of AI agents to track elapsed wall-clock time [6, 7] motivate the question, but whether a hierarchy of internal rhythms would remedy them is a further extrapolation these tests do not address.

## Competing explanations

First, oscillations might be epiphenomenal, and the objection comes in two forms. In the strong form they are a causally inert readout of computation; behavioral rhythmicity under aperiodic input counts against it, to the extent those effects hold. In the weak form they are causally active but functionally incidental, a by-product of excitation–inhibition dynamics that modulates excitability without serving computation [102], where inter-areal coherence merely reflects underlying communication [103]. Behavioral rhythmicity cannot settle the weak form, since a secondary process that periodically lowers excitability would leave the same behavioral signature. The decisive test is interventional: in a model that reproduces behavioral rhythmicity, clamp or remove the rhythm while holding inputs and parameters fixed, and ask whether task performance falls or rises.

Second, several architectures already model temporally structured information, so any test of the hypothesis should include them. *Reservoir computing* is a well-established example: it projects recent input history into a high-dimensional recurrent reservoir, whose dynamics are then read out for the task [104]. State-space models maintain a latent state that is updated across sequence steps, allowing them to capture long-range dependencies [21, 46]. Recurrent neural networks can also learn oscillatory regimes when trained on temporal tasks [29]. Related approaches, ranging from dynamical-systems reconstruction [105, 106] to *recurrent switching linear dynamical systems* [107], neural ordinary differential equations [25] learn temporal dynamics directly from data rather than imposing an oscillatory structure.

Third, for phase-dependent routing and cross-frequency multiplexing, alternative architectures may reproduce aspects of the proposed computational functions without explicitly implementing oscillatory mechanisms. For instance, recurrent neural networks can learn controllable activity slots that support sequence memory without an imposed theta–gamma code [108]. Transformers, meanwhile, combine content representations with rotary positional embeddings, where position-dependent phase rotations encode relative sequence position and influence attention [3, 4]. However, these phases are indexed by sequence position rather than generated by autonomous temporal dynamics. Whether an internally generated temporal reference provides additional computational advantages beyond learned recurrent dynamics and position-indexed attention therefore remains an open question. The key test is whether an oscillatory phase variable, as a computational primitive, improves task performance or generalization, under matched-resource model

For delay-structured representation, standard architectures do not treat connection-specific delays as explicit computational variables [109], so a natural test is not a direct comparison. This makes the delay-shuffle test particularly diagnostic (Fig. 4). For offline consolidation, the functional alternative is prioritized experience replay, which already selects what to replay by its value for learning [99, 100]. The hypothesis must therefore show what rhythm-gated timing adds once replay content and count are matched.

## Three research directions

How can endogenous temporal structure be implemented in AI architectures? The four mechanisms identified motivate several approaches, some based on biological mechanisms and others abstracting them as intrinsic state variables. We outline three directions, each with a testable prediction, and specify the requirements for a fair comparison in Box 3.

**Oscillatory memory systems.** Maintain a dynamical state between processing steps, organized around stable oscillatory regimes, with information phase-coded and selectively gated by the current oscillatory context. Unlike standard recurrent networks, which struggle to hold information across multiple timescales [1], oscillatory systems separate timescales through their frequency structure. *Testable prediction*: superior performance on tasks that require holding several items and retrieving them in a specific order (the operation of theta–gamma neural code), and equivalent performance on unordered or single-item tasks.

**Phase-gated routing architectures.** Replace or supplement attention with routing rules that depend on the phase relationship between modules, giving state-dependent connectivity without enumerating configurations. Unlike attention, which recomputes routing from the input content at each step, phase-based routing reconfigures through the ongoing dynamics. *Testable prediction:* performance equivalent to attention with substantially fewer parameters in tasks requiring rapid reconfiguration of information flow, such as multi-task settings where one network must route differently across tasks.

**Rhythmic predictive coding.** Extend predictive-processing architectures [110] so that the prediction itself is timed: instead of forecasting only what the next input will be, the system also predicts when it will arrive, using an ongoing oscillatory phase. Unlike models that treat inputs as arriving on demand, such a system carries an internal expectation of when the next input is due. *Testable prediction*: improved robustness to temporal noise and jitter in dynamic real-time environments, where current AI models exhibit temporal blindness [6, 7].

**Box 3. A diagnostic test for endogenous temporal structure**

Each research direction above predicts an advantage on particular tasks, but such advantages are hard to attribute. A model with oscillatory dynamics may outperform a baseline for reasons unrelated to the oscillation itself, including extra parameters, longer training, or a better-tuned optimizer. Making an advantage traceable to the temporal mechanism itself rests on three conditions:

**The comparison must be fair.** The oscillatory model should be compared with mainstream architectures and matched as closely as possible for parameter count, data size, training compute, and inference cost. Reported performance should include accuracy, memory use, run-time, and stability under temporal perturbations. Exact equality across architectures is challenging, but these controls make it more likely that an apparent gain reflects the mechanism itself rather than extra capacity, longer training, or higher compute.

**The mechanism must be ablated.** For each function, the putative causal variable is disrupted while other factors are held fixed. For delay-based capacity, the test compares learned delays with zero and shuffled controls; for phase-dependent routing, the test uses random-phase controls; for oscillatory multiplexing, the test preserves the encoded items while scrambling, jittering, or removing the phase-slot structure; and for offline consolidation, the test delivers the same replay content on a fixed or random schedule instead of gating it by coupled internal rhythms. The predicted advantage should be reduced or abolished: the temporal variable must confer the advantage when intact and fail when disrupted.

**The prediction must be falsifiable.** Before the experiment is run, the task families expected to benefit and the magnitude and direction of the predicted effect are specified. If oscillatory dynamics improve all tasks equally, or none of them, the specific mechanism claim is weakened. A bounded negative result is informative when oscillation does not yield the predicted benefit in a defined setting, because it maps the limits of the mechanism. This keeps the hypothesis falsifiable: endogenous temporal structure should confer selective advantages in the predicted regimes, not uniform gains across tasks.

The **delay-ablation test** meets these criteria. (1) The comparison is fair: capacity and memory are evaluated separately across chain lengths. (2) The mechanism is selectively disrupted, by shuffling delay while holding synaptic weights fixed. (3) Delay shuffling substantially impairs retrieval of sequences, particularly for longer chains. In contrast, eliminating delays collapses capacity itself, while abolishing recall only for longer chains.

Together, these conditions make the delay-ablation result mechanistically interpretable. However, demonstrating the functional contribution of a temporal mechanism does not establish that its endogenous generation is necessary. For oscillatory mechanisms, establishing this requires a further control: comparing a model whose oscillations arise from its own internal dynamics with one that receives an equivalent oscillatory signal externally. A performance advantage in the self-generated condition would indicate a specific computational contribution of endogenous generation; comparable performance would instead suggest that the benefit comes from temporal structure itself, rather than from generating it internally. What is missing is not the tooling to build such systems, but a disciplined comparison under fair, matched, and pre-specified conditions.

The core proposition of this article is that endogenous temporal structure—internally generated rhythms, phase relationships, and connection-specific delays—offers computational advantages that are not explicitly implemented in mainstream reactive, input-driven architectures. The delay-ablation result provides one concrete example, and analogous functional contributions can be tested directly for the other three mechanisms. For example, phase relationships can be randomized to test phase-dependent routing or cross-frequency multiplexing, while oscillatory timing can be perturbed to test oscillation-gated consolidation. The central question, therefore, is not whether AI needs biological clocks, but under which conditions the internally generated temporal structure confers a computational advantage. Beyond the four specific functions identified in this article, a broader possibility is that endogenous temporal structure could provide a reusable temporal framework: a self-maintained internal reference through which multiple processes coordinate their activity as computational demands change, rather than having each timing rule specified externally via the input. Whether such an internally maintained temporal reference could address practical limitations, such as those observed in contemporary AI agents, remains to be investigated.

Ultimately, self-generated rhythms or internal timing relations are not themselves the test; their computational relevance lies in whether they alter how information is routed, stored, or combined. An internal temporal reference becomes computationally operative when its evolving state modulates how information is processed, even when the input is held fixed. Establishing their functional contribution—through matched-resource, mechanism-specific comparisons—would move endogenous temporal structure from a set of biological analogies to a design principle for future sequence-processing architectures.

## Conclusion

The absence of clocks in sequence-processing AI is not a minor technical detail. It reflects an architectural choice: to embed time in the input stream rather than as an internal computational variable. For many tasks this choice is sufficient and its costs are rarely exposed. However, for those requiring temporal reasoning, continual adaptation, and dynamic coordination under variable physical timing, the consequences of this choice may become a fundamental constraint.

Neuroscience has spent decades studying how internal brain rhythms that occur without external stimuli support cognitive functions. The computational principles that have been identified—phase-dependent routing, delay-structured representation, cross-frequency multiplexing, and oscillation-gated consolidation—are not incidental biological details. They are candidate solutions to computational problems that any general-purpose system operating in temporal settings must confront. Whether sequence models need to implement these principles directly, approximate them, or find alternative solutions to the same problems is an empirical question. Crucially, these claims are testable. The delay-ablation experiment is illustrative: shuffling delays disrupts retrieval of learned sequences more strongly than it reduces capacity, especially for sequences propagating deeper through the network. Removing delays entirely, as opposed to permuting them, collapses capacity itself, with retrieval of longer sequences especially vulnerable. Delay structure therefore plays two complementary roles: supporting a combinatorial repertoire, and more selectively, determining which learned sequences within that repertoire can be reliably retrieved. The same diagnostic logic extends to the other three principles. What is not in question is that the problems exist, that several biological mechanisms are increasingly well characterized, and that the field now has tractable ways to test whether they provide useful computational advantages in artificial systems (see Open Questions).

## Open Questions

- Do self-generated oscillations confer computational advantages or can sequence models approximate their functions under temporal perturbations?
- Can artificial networks learn and adapt intrinsic temporal structure to changing task demands without destabilizing the dynamics required for computation?
- Can oscillatory phase serve as a computational variable for selective retrieval, routing, and sequence transformation within a shared recurrent state?
- Can phase-dependent routing match attention mechanisms with fewer parameters under dynamically changing, context-dependent demands?
- Does consolidation gated by internal rhythms improve retention or sample efficiency over the same replay delivered on a fixed or random schedule?
- Do nested oscillations in artificial systems reflect general computational principles, or is such a frequency hierarchy specific to biological constraints?

## Glossary

**Attention mechanism:** a neural-network operation that computes context-dependent weights over a set of input representations, determining how much each element contributes to the update. In transformers, these weights are computed at each layer from learned projections of the input itself, allowing the network to selectively combine information across the entire sequence.

**Autoregressive generation**: a sequential method in which a model generates one element at a time, feeding each emitted element back as input for the next step. The model can run without external input, but its output still unfolds one element per step.

**Communication through coherence:** the hypothesis that effective connectivity between neural populations depends not only on anatomical connections and synaptic weights but also on the phase relationship between their oscillations. Phase-aligned regions exchange information more efficiently, while phase-misaligned regions communicate less effectively.

**Complementary learning systems:** the proposal that the brain pairs a fast-learning hippocampal store with a slow-learning neocortical system, and that interleaved replay from the first trains the second without overwriting what it already knows.

**Cross-frequency coupling:** a statistical dependence between rhythms of different frequencies, most often via phase–amplitude coupling as a candidate mechanism for temporal multiplexing, in which the phase of a slower rhythm modulates the amplitude of a faster rhythm.

**Delay differential equation:** a mathematical formulation for describing systems in which temporal evolution depends not only on the present state, but also on past states at fixed or distributed lags. In neuroscience, it is used to model network dynamics shaped by conduction delays.

**Endogenous temporal structure:** patterns such as rhythm, phase, and delay that are generated or maintained within a system and are available to shape its own computation, as opposed to structure supplied only by the ordering of external input.

**Experience replay**: a machine-learning method in which stored past experiences are held in a replay buffer and sampled during training. This improves sample efficiency (reusing each stored experience across many updates), mitigates forgetting (revisiting older examples during training), and stabilizes learning (breaking correlations between consecutive updates). Samples can be drawn uniformly or prioritized by how much each is expected to improve learning.

**Mixture-of-experts:** a neural network architecture in which a learned router (a small trained network) decides which of several specialized sub-networks, called experts, should process each input. It provides sparse, input-dependent routing, but not necessarily phase-dependent routing.

**Neural ordinary differential equation:** an alternative to recurrent architectures that efficiently approximates autonomous nonlinear dynamics with continuous-time hidden-state evolution, adapts its evaluation strategy to each input, and has constant memory cost.

**Polychronous neuronal group:** a precise but asynchronous firing cascade that emerges spontaneously in recurrent spiking networks with synaptic plasticity. Heterogeneous delays allow spikes emitted at different times to converge on the same postsynaptic neurons; because each neuron can participate in many distinct patterns at different relative timings, the count of coexisting groups can scale combinatorially, providing a rich substrate for memory and computation.

**Positional encoding:** additional information supplied to a sequence model to indicate the order of its inputs. In transformers, position is encoded either by a fixed function of the input's index or by a learned vector associated with each position during training.

**Recurrent neural network:** a sequential architecture whose output at each step depends on both the current input and an updated hidden state that maintains a summary of the past. Extensions such as long short-term memory and gated recurrent units use mechanisms that regulate how information is retained, updated, or discarded across time steps.

**Recurrent switching linear dynamical system:** a state-space model that approximates nonlinear dynamics as a sequence of discrete states, each governed by its own linear dynamics, with a learned transition between states over time. It is often used to decompose neural data into interpretable latent states, with switching that depends on the recent activity and behavioral data.

**Reservoir computing:** a computational framework in which inputs drive a fixed, high-dimensional recurrent network, and only a readout layer is trained. The recurrent network is initialized with random connections and left untrained; its transient responses to input act as a rich set of nonlinear features that a simple linear readout can then combine to produce the output.

**Retrieval-augmented generation:** a technique that enhances large language models by fetching relevant information from an external data source at inference time and conditioning the response on that content. Retrieval is typically triggered by the input query and performed by a separate module, without modifying the model's trained parameters.

**Sharp-wave ripples:** brief hippocampal population events, carrying a fast oscillation at 110–200 Hz in rodents and around 80–100 Hz in humans. They occur during non-REM sleep and quiet wakefulness and carry compressed replay of past experience and, during awake rest, sequences along possible future paths. During sleep they are nested within spindles and slow oscillations.

**Spike-timing-dependent plasticity:** a local synaptic learning rule in which the sign and magnitude of synaptic change depend on the relative timing of pre- and post-synaptic activity. In its classical form, a synapse is strengthened when presynaptic activity precedes postsynaptic activity, and weakened when the order is reversed.

**State-space model:** a sequence-processing architecture, rooted in control theory, that maintains a latent state updated by each input. The state evolves according to a linear recurrence parameterized by a small set of matrices; modern instantiations make these matrices input-dependent, allowing the model to selectively retain or forget past information.

**Theta–gamma neural code:** a proposed mechanism in which several gamma subcycles (30–80 Hz) are nested within a single theta cycle (4–8 Hz), with each gamma subcycle representing a different item (e.g., a memory, stimulus, or location). The active neurons indicate item identity (which one it is), and the position of gamma cycles indicates temporal order (when it occurs).

**Transformer:** a non-recurrent architecture that has become the dominant model for sequential modeling such as natural language processing. It processes ordered inputs in parallel using self-attention (which mixes information across positions) and positional encodings—either added at the input or injected within the attention mechanism—to convey order.

**Transmission delay:** the time required for a signal to travel from one point to another due to finite propagation speed. In neuroscience, it refers to the time taken for an action potential to travel along an axon, ranging from about sub-milliseconds to tens of milliseconds. Delays depend on axon diameter and myelination, and are heterogeneous in the brain and across mammalian species.

## Acknowledgment:

M.H. thanks Viktor Sip, Marmaduke Woodman, and Spase Petkoski for fruitful discussions. This research has received funding from the European Union's Horizon Europe Programme under the Specific Grant Agreement No. 101147319 (EBRAINS 2.0 Project) to M.H. and V.J., No. 101137289 (Virtual Brain Twin Project) to V.J., and government grant by the Agence Nationale de la Recherche reference ANR-22-PESN-0012 (France 2030 program) to V.J.

## Data availability:

No new empirical data were collected in this study. Simulation data were generated as part of the analyses described here. Code for reproducing the simulations and analyses is available on GitHub (https://github.com/mhashemi0873/Clocks_AI).

## Declaration of Competing Interest

The authors declare that they have no known competing financial interests or personal relationships that could have appeared to influence the work reported in this paper.

## References

1. Tiezzi, M. et al. (2025) Back to recurrent processing at the crossroad of transformers and state-space models. Nat. Mach. Intell. 7, 678–688.
2. Vaswani, A. et al. (2017) Attention is all you need. In Advances in Neural Information Processing Systems 30, 5998–6008.
3. Dufter, P. et al. (2022) Position information in transformers: an overview. Comput. Linguist. 48, 733–763.
4. Su, J. et al. (2024) RoFormer: enhanced transformer with rotary position embedding. Neurocomputing 568, 127063.
5. Wang, L. et al. (2024) A survey on large language model based autonomous agents. Front. Comput. Sci. 18, 186345.
6. Cheng, Y. et al. (2026) Your LLM agents are temporally blind: the misalignment between tool use decisions and human time perception. In Findings of the Association for Computational Linguistics: ACL 2026, 37082–37104.
7. Sehgal, N. et al. (2026) Real-time deadlines reveal fragile temporal adaptation in LLM strategic dialogues. In Proceedings of the 2026 Conference on Empirical Methods in Natural Language Processing, 17735–17756.

8. Buonomano, D.V. and Maass, W. (2009) State-dependent computations: spatiotemporal processing in cortical networks. Nat. Rev. Neurosci. 10, 113–125.
9. Eichenbaum, H. (2014) Time cells in the hippocampus: a new dimension for mapping memories. Nat. Rev. Neurosci. 15, 732–744.
10. Mattar, M.G. and Lengyel, M. (2022) Planning in the brain. Neuron 110, 914–934.
11. Fries, P. (2015) Rhythms for cognition: communication through coherence. Neuron 88, 220–235.
12. Gwilliams, L. (2026) Time after time: neural codes of speech comprehension. Trends Cogn. Sci. (online ahead of print).
13. Buzsáki, G. and Draguhn, A. (2004) Neuronal oscillations in cortical networks. Science 304, 1926–1929.
14. Buzsáki, G. (2026) Time, space, memory and brain–body rhythms. Nat. Rev. Neurosci. 27, 61–78.
15. Lisman, J.E. and Jensen, O. (2013) The theta-gamma neural code. Neuron 77, 1002–1016.
16. Engel, A.K. et al. (2001) Dynamic predictions: oscillations and synchrony in top–down processing. Nat. Rev. Neurosci. 2, 704–716.
17. Hyafil, A. et al. (2015) Neural cross-frequency coupling: connecting architectures, mechanisms, and functions. Trends Neurosci. 38, 725–740.
18. Feldman, J. (2013) The neural binding problem(s). Cogn. Neurodyn. 7, 1–11.
19. Singer, W. (1999) Neuronal synchrony: a versatile code for the definition of relations? Neuron 24, 49–65.
20. Fries, P. (2005) A mechanism for cognitive dynamics: neuronal communication through neuronal coherence. Trends Cogn. Sci. 9, 474–480.
21. Gu, A. et al. (2022) Efficiently modeling long sequences with structured state spaces. In International Conference on Learning Representations.
22. Wilson, H.R. and Cowan, J.D. (1972) Excitatory and inhibitory interactions in localized populations of model neurons. Biophys. J. 12, 1–24.
23. Petkoski, S. and Jirsa, V.K. (2019) Transmission time delays organize the brain network synchronization. Philos. Trans. R. Soc. A 377, 20180132.
24. Pastalkova, E. et al. (2008) Internally generated cell assembly sequences in the rat hippocampus. Science 321, 1322–1327.
25. Chen, R.T.Q. et al. (2018) Neural ordinary differential equations. In Advances in Neural Information Processing Systems 31, 6571–6583.
26. Ságodi, Á. and Park, I.M. (2026) Universal approximation theorems for dynamical systems with infinite-time horizon guarantees. In Advances in Neural Information Processing Systems (NeurIPS 2026; accepted).
27. Effenberger, F. et al. (2025) The functional role of oscillatory dynamics in neocortical circuits: a computational perspective. Proc. Natl. Acad. Sci. U. S. A. 122, e2412830122.
28. Rusch, T.K. and Rus, D. (2025) Oscillatory state-space models. In International Conference on Learning Representations, 22303–22322.
29. Pals, M. et al. (2024) Trained recurrent neural networks develop phase-locked limit cycles in a working memory task. PLoS Comput. Biol. 20, e1011852.
30. VanRullen, R. (2016) Perceptual cycles. Trends Cogn. Sci. 20, 723–735.

31. Busch, N.A. et al. (2009) The phase of ongoing EEG oscillations predicts visual perception. J. Neurosci. 29, 7869–7876.
32. Mathewson, K.E. et al. (2009) To see or not to see: prestimulus α phase predicts visual awareness. J. Neurosci. 29, 2725–2732.
33. VanRullen, R. et al. (2007) The blinking spotlight of attention. Proc. Natl. Acad. Sci. U. S. A. 104, 19204–19209.
34. Landau, A.N. and Fries, P. (2012) Attention samples stimuli rhythmically. Curr. Biol. 22, 1000–1004.
35. Helfrich, R.F. et al. (2018) Neural mechanisms of sustained attention are rhythmic. Neuron 99, 854–865.e5.
36. Ideriha, T. and Ushiyama, J. (2024) Behavioral fluctuation reflecting theta-rhythmic activation of sequential working memory. Sci. Rep. 14, 550.
37. Fiebelkorn, I.C. et al. (2011) Ready, set, reset: stimulus-locked periodicity in behavioral performance demonstrates the consequences of cross-sensory phase reset. J. Neurosci. 31, 9971–9981.
38. Ruzzoli, M. et al. (2019) The relevance of alpha phase in human perception. Cortex 120, 249–268.
39. Brookshire, G. (2022) Putative rhythms in attentional switching can be explained by aperiodic temporal structure. Nat. Hum. Behav. 6, 1280–1291.
40. Samaha, J. and Postle, B.R. (2015) The speed of alpha-band oscillations predicts the temporal resolution of visual perception. Curr. Biol. 25, 2985–2990.
41. Buergers, S. and Noppeney, U. (2022) The role of alpha oscillations in temporal binding within and across the senses. Nat. Hum. Behav. 6, 732–742.
42. Fiebelkorn, I.C. and Kastner, S. (2019) A rhythmic theory of attention. Trends Cogn. Sci. 23, 87–101.
43. Bahdanau, D. et al. (2015) Neural machine translation by jointly learning to align and translate. In International Conference on Learning Representations.
44. Elman, J.L. (1990) Finding structure in time. Cogn. Sci. 14, 179–211.
45. Shazeer, N. et al. (2017) Outrageously large neural networks: the sparsely-gated mixture-of-experts layer. In International Conference on Learning Representations.
46. Dao, T. and Gu, A. (2024) Transformers are SSMs: generalized models and efficient algorithms through structured state space duality. In Proceedings of the 41st International Conference on Machine Learning, Proc. Mach. Learn. Res. 235, 10041–10071.
47. Akam, T. and Kullmann, D.M. (2010) Oscillations and filtering networks support flexible routing of information. Neuron 67, 308–320.
48. Pariz, A. et al. (2021) Transmission delays and frequency detuning can regulate information flow between brain regions. PLoS Comput. Biol. 17, e1008129.
49. Palmigiano, A. et al. (2017) Flexible information routing by transient synchrony. Nat. Neurosci. 20, 1014–1022.
50. Womelsdorf, T. et al. (2007) Modulation of neuronal interactions through neuronal synchronization. Science 316, 1609–1612.
51. Bastos, A.M. et al. (2015) Visual areas exert feedforward and feedback influences through distinct frequency channels. Neuron 85, 390–401.

52. Bonnefond, M. et al. (2017) Communication between brain areas based on nested oscillations. eNeuro 4, ENEURO.0153-16.2017.
53. Jensen, O. and Mazaheri, A. (2010) Shaping functional architecture by oscillatory alpha activity: gating by inhibition. Front. Hum. Neurosci. 4, 186.
54. Lakatos, P. et al. (2008) Entrainment of neuronal oscillations as a mechanism of attentional selection. Science 320, 110–113.
55. Salami, M. et al. (2003) Change of conduction velocity by regional myelination yields constant latency irrespective of distance between thalamus and cortex. Proc. Natl. Acad. Sci. U. S. A. 100, 6174–6179.
56. Hashemi, M. et al. (2017) Anesthetic action on the transmission delay between cortex and thalamus explains the beta-buzz observed under propofol anesthesia. PLoS One 12, e0179286.
57. Caminiti, R. et al. (2013) Diameter, length, speed, and conduction delay of callosal axons in macaque monkeys and humans: comparing data from histology and magnetic resonance imaging diffusion tractography. J. Neurosci. 33, 14501–14511.
58. Hashemi, M. et al. (2012) Effect of duration of synaptic activity on spike rate of a Hodgkin-Huxley neuron with delayed feedback. Phys. Rev. E 85, 021917.
59. Lefebvre, J. et al. (2025) Myelin-induced gain control in nonlinear neural networks. Commun. Phys. 8, 145.
60. Jirsa, V.K. and Haken, H. (1996) Field theory of electromagnetic brain activity. Phys. Rev. Lett. 77, 960–963.
61. Jirsa, V.K. (2009) Neural field dynamics with local and global connectivity and time delay. Philos. Trans. R. Soc. A 367, 1131–1143.
62. Deco, G. et al. (2009) Key role of coupling, delay, and noise in resting brain fluctuations. Proc. Natl. Acad. Sci. U. S. A. 106, 10302–10307.
63. Hashemi, M. et al. (2026) Principles and operation of virtual brain twins. IEEE Rev. Biomed. Eng. 19, 111–139.
64. Vicente, R. et al. (2008) Dynamical relaying can yield zero time lag neuronal synchrony despite long conduction delays. Proc. Natl. Acad. Sci. U. S. A. 105, 17157–17162.
65. Petkoski, S. and Jirsa, V.K. (2022) Normalizing the brain connectome for communication through synchronization. Netw. Neurosci. 6, 722–744.
66. Yeung, M.K.S. and Strogatz, S.H. (1999) Time delay in the Kuramoto model of coupled oscillators. Phys. Rev. Lett. 82, 648–651.
67. Madadi Asl, M. et al. (2018) Dendritic and axonal propagation delays may shape neuronal networks with plastic synapses. Front. Physiol. 9, 1849.
68. Ghadiri, A. et al. (2026) Delay selection by spike-timing-dependent plasticity shapes efficient networks for signal transmission. J. Neurosci. 46, e0103262026.
69. Izhikevich, E.M. (2006) Polychronization: computation with spikes. Neural Comput. 18, 245–282.
70. Otto, A. et al. (2019) Nonlinear dynamics of delay systems: an overview. Philos. Trans. R. Soc. A 377, 20180389.
71. Bastos, A.M. et al. (2015) Communication through coherence with inter-areal delays. Curr. Opin. Neurobiol. 31, 173–180.

72. Breakspear, M. (2017) Dynamic models of large-scale brain activity. Nat. Neurosci. 20, 340–352.

73. Cabral, J. et al. (2022) Metastable oscillatory modes emerge from synchronization in the brain spacetime connectome. Commun. Phys. 5, 184.

74. Hopfield, J.J. (1982) Neural networks and physical systems with emergent collective computational abilities. Proc. Natl. Acad. Sci. U. S. A. 79, 2554–2558.

75. Sompolinsky, H. and Kanter, I. (1986) Temporal association in asymmetric neural networks. Phys. Rev. Lett. 57, 2861–2864.

76. Punetha, N. et al. (2015) Bipartite networks of oscillators with distributed delays: synchronization branches and multistability. Phys. Rev. E 91, 042906.

77. Hammouamri, I. et al. (2024) Learning delays in spiking neural networks using dilated convolutions with learnable spacings. In International Conference on Learning Representations, 17890–17903.

78. Aru, J. et al. (2015) Untangling cross-frequency coupling in neuroscience. Curr. Opin. Neurobiol. 31, 51–61.

79. McLelland, D. and VanRullen, R. (2016) Theta-gamma coding meets communication-through-coherence: neuronal oscillatory multiplexing theories reconciled. PLoS Comput. Biol. 12, e1005162.

80. Akam, T. and Kullmann, D.M. (2014) Oscillatory multiplexing of population codes for selective communication in the mammalian brain. Nat. Rev. Neurosci. 15, 111–122.

81. Lisman, J. (2005) The theta/gamma discrete phase code occuring during the hippocampal phase precession may be a more general brain coding scheme. Hippocampus 15, 913–922.

82. Lisman, J.E. and Idiart, M.A.P. (1995) Storage of 7 ± 2 short-term memories in oscillatory subcycles. Science 267, 1512–1515.

83. Bahramisharif, A. et al. (2018) Serial representation of items during working memory maintenance at letter-selective cortical sites. PLoS Biol. 16, e2003805.

84. Axmacher, N. et al. (2010) Cross-frequency coupling supports multi-item working memory in the human hippocampus. Proc. Natl. Acad. Sci. U. S. A. 107, 3228–3233.

85. Barr, M.S. et al. (2017) Impaired theta-gamma coupling during working memory performance in schizophrenia. Schizophr. Res. 189, 104–110.

86. Karlsson, A.E. et al. (2022) Out of rhythm: compromised precision of theta-gamma coupling impairs associative memory in old age. J. Neurosci. 42, 1752–1764.

87. Duecker, K. et al. (2024) Oscillations in an artificial neural network convert competing inputs into a temporal code. PLoS Comput. Biol. 20, e1012429.

88. Staresina, B.P. et al. (2023) How coupled slow oscillations, spindles and ripples coordinate neuronal processing and communication during human sleep. Nat. Neurosci. 26, 1429–1437.

89. Buzsáki, G. (2015) Hippocampal sharp wave-ripple: a cognitive biomarker for episodic memory and planning. Hippocampus 25, 1073–1188.

90. Staresina, B.P. et al. (2015) Hierarchical nesting of slow oscillations, spindles and ripples in the human hippocampus during sleep. Nat. Neurosci. 18, 1679–1686.

91. Girardeau, G. et al. (2009) Selective suppression of hippocampal ripples impairs spatial memory. Nat. Neurosci. 12, 1222–1223.

92. Pfeiffer, B.E. and Foster, D.J. (2013) Hippocampal place-cell sequences depict future paths to remembered goals. Nature 497, 74–79.
93. Ólafsdóttir, H.F. et al. (2015) Hippocampal place cells construct reward related sequences through unexplored space. eLife 4, e06063.
94. Maingret, N. et al. (2016) Hippocampo-cortical coupling mediates memory consolidation during sleep. Nat. Neurosci. 19, 959–964.
95. Fernández-Ruiz, A. et al. (2019) Long-duration hippocampal sharp wave ripples improve memory. Science 364, 1082–1086.
96. Lewis, P. et al. (2020) Retrieval-augmented generation for knowledge-intensive NLP tasks. In Advances in Neural Information Processing Systems 33, 9459–9474.
97. Parisi, G.I. et al. (2019) Continual lifelong learning with neural networks: a review. Neural Netw. 113, 54–71.
98. van de Ven, G.M. et al. (2020) Brain-inspired replay for continual learning with artificial neural networks. Nat. Commun. 11, 4069.
99. Schaul, T. et al. (2016) Prioritized experience replay. In International Conference on Learning Representations.
100. Mattar, M.G. and Daw, N.D. (2018) Prioritized memory access explains planning and hippocampal replay. Nat. Neurosci. 21, 1609–1617.
101. McClelland, J.L. et al. (1995) Why there are complementary learning systems in the hippocampus and neocortex: insights from the successes and failures of connectionist models of learning and memory. Psychol. Rev. 102, 419–457.
102. Ray, S. and Maunsell, J.H.R. (2015) Do gamma oscillations play a role in cerebral cortex? Trends Cogn. Sci. 19, 78–85.
103. Schneider, M. et al. (2021) A mechanism for inter-areal coherence through communication based on connectivity and oscillatory power. Neuron 109, 4050–4067.e12.
104. Maass, W. et al. (2002) Real-time computing without stable states: a new framework for neural computation based on perturbations. Neural Comput. 14, 2531–2560.
105. Durstewitz, D. et al. (2023) Reconstructing computational system dynamics from neural data with recurrent neural networks. Nat. Rev. Neurosci. 24, 693–710.
106. Sip, V. et al. (2023) Characterization of regional differences in resting-state fMRI with a data-driven network model of brain dynamics. Sci. Adv. 9, eabq7547.
107. Linderman, S. et al. (2017) Bayesian learning and inference in recurrent switching linear dynamical systems. In Proceedings of the 20th International Conference on Artificial Intelligence and Statistics, Proc. Mach. Learn. Res. 54, 914–922.
108. Whittington, J.C.R. et al. (2025) A tale of two algorithms: structured slots explain prefrontal sequence memory and are unified with hippocampal cognitive maps. Neuron 113, 321–333.e6.
109. Izhikevich, E.M. (2025) Spiking Manifesto. arXiv:2512.11843 (preprint).
110. Rao, R.P.N. and Ballard, D.H. (1999) Predictive coding in the visual cortex: a functional interpretation of some extra-classical receptive-field effects. Nat. Neurosci. 2, 79–87.

## Supplementary Methods

### 1. Communication through coherence

We simulated three signals, a sender $X(t)$, a competitor $Y(t)$, and a receiver $Z(t)$, at $f_s = 2000$ Hz for $T = 120$ sec. A slowly varying message signal was defined as $m(t) = 1 + m \cdot LP_3[\eta(t)]$, where $\eta(t)$ is i.i.d. standard Gaussian noise, $LP_3$ a 3rd-order Butterworth low-pass filter with a cut-off at 4 Hz, and $m = 0.5$ the depth controlling the amplitude of the slow fluctuations. Signals $X$ and $Z$ carried the same message, delayed by $\delta = 30$ ms in $Z$, whereas $Y$ carried an independent message except in the alpha-gating. Independent Gaussian noise $\varepsilon(t) \sim \mathcal{N}(0, \sigma^2)$, with $\sigma = 0.02$ was added to each signal.

### 1.1 Gamma synchronization

In this regime, the three signals were defined as follows:

$$X(t) = A_\gamma m(t) sin\left(2\pi f_\gamma t\right) + \varepsilon_X(t)$$
$$Y(t) = A_\gamma m'(t) sin\left(2\pi f_\gamma t + \pi\right) + \varepsilon_Y(t)$$
$$Z(t) = A_\gamma m(t - \delta) sin\left(2\pi f_\gamma t\right) + \varepsilon_Z(t)$$

with $f_\gamma = 80$ Hz, and $A_\gamma = 2.0$. The signal pair $X$ and $Z$ carried the same underlying message signal and were phase-aligned, whereas $Y$ carried an independent message and was antiphase relative to both $X$ and $Z$. The time delay $\delta$ was introduced only to $m(t)$ in $Z$, leaving its carrier phase-locked with $X$ during the simulation.

### 1.2 Alpha gating

In this regime, the three signals were defined as follows:

$$X(t) = A_X m(t) sin\left(2\pi f_\alpha t\right) + \varepsilon_X(t)$$
$$Y(t) = A_Y sin\left(2\pi f_\alpha t\right) + \varepsilon_Y(t)$$
$$Z(t) = A_Z m(t - \delta) sin\left(2\pi f_\alpha t\right) + \varepsilon_Z(t),$$

with $f_{\alpha} = 10$ Hz, $A_X = A_Z = 2.0$, and $A_Y = 4.0$. All three oscillations were in phase, whereas $Y$ had a larger carrier amplitude than $X$ and $Z$, but no message modulation.

### 1.3 Nested alpha–gamma activity

The third regime combined an alpha oscillation at $f_{\alpha} = 10$ Hz with gamma bursts at $f_{\gamma} = 80$ Hz. To simplify the notation, $\phi(t) = 2\pi f_{\alpha} t$ represents the alpha phase and $\psi(t) = 2\pi f_{\gamma} t$ denotes the gamma phase. The signals were defined as follows:

$$X(t) = A_{\alpha} sin[\phi(t)] + A_{\gamma,X} g[\phi(t)] m(t) sin[\psi(t)] + \varepsilon_X(t)$$
$$Y(t) = A_Y sin[\phi(t) + \pi] + A_{\gamma,Y} g[\phi(t) + \pi] m'(t) sin[\psi(t)] + \varepsilon_Y(t)$$
$$Z(t) = A_{\alpha} sin[\phi(t)] + A_{\gamma,X} g[\phi(t)] m(t - \delta) sin[\psi(t)] + \varepsilon_Z(t)$$

with $A_{\alpha} = 2.0$, $A_Y = 4.0$, $A_{\gamma,X} = 1.0$ and $A_{\gamma,Y} = 0.6$. The gamma bursts in $X$ and $Z$ occurred at the same alpha phase and carried the shared message. Signal $Y$ had a stronger alpha oscillation that was out of phase with $X$ and $Z$; its smaller gamma bursts followed this shifted alpha phase and carried an independent message. The gating function was $g(\phi) = 1$ when $\left|\phi - \phi_{peak}\right| \leq n_{burst}\pi/8$, and $n_{burst} = 5$. Out of this window, the function dropped to zero over a cosine ramp of 0.45 rad, and remained zero elsewhere.

### 1.4 Coupling metrics

We calculated four coupling metrics for the $X$–$Z$ and $Y$–$Z$ signal pairs, as follows.

**Coherence**. For a pair $X$ and $Y$, the magnitude-squared coherence was defined as

$$C_{xy}(f) = \frac{\left|S_{xy}(f)\right|^2}{S_{xx}(f) \cdot S_{yy}(f)},$$

where $S_{xy}(f)$ is the cross-spectral density while $S_{xx}(f)$ and $S_{yy}(f)$ are the auto-spectral densities. We used Welch's method with a segment length of $f_s = 2000$ samples to achieve a 1 Hz resolution. Coherence was then averaged across the relevant frequency band. This metric can be high for both in-phase and out-of-phase oscillations.

**Envelope correlation**. We band-pass filtered each signal (4th-order Butterworth) in the frequency band of interest. The envelope $e(t) = |a(t)|$ was extracted from the analytic signal $a(t)$ using the Hilbert transform. We discarded the first and last 400 samples (200 ms at each end) to reduce transients, then calculated the Pearson correlation between each pair.

**Time-delayed mutual information**. We low-pass filtered the envelopes at 5 Hz, downsampled tenfold, standardized (zero mean, unit variance), and then discretized into $n_{bins} = 8$ uniform bins. For the signals $X$ and $Z$, mutual information was defined as:

$$I(X; Z) = \sum_{x,z} p(x, z)\, log\left[\frac{p(x,z)}{p(x)p(z)}\right],$$

where $p(x, z)$ is the joint probability of binned values, while $p(x)$ and $p(z)$ are the individual marginal probabilities. We reported the maximum mutual information found across time delays of 1–14 samples after downsampling (5–70 ms).

**Transfer entropy**. To measure information flow from $X$ to $Z$, we evaluated the source envelope at a lag of six samples before the receiver (30 ms at the downsampled rate of 200 Hz, matching the simulated message delay), and calculated

$$TE(X\rightarrow Z) = H\left(Z_{t+1}, Z_t\right) + H\left(X_t, Z_t\right) - H\left(Z_t\right) - H\left(Z_{t+1}, X_t, Z_t\right),$$

where $H$ is Shannon entropy estimated from bin counts. This metric indicates directed statistical dependence but it does not establish causality on its own.

## 2. Polychronous neuronal groups

To investigate how transmission delays support the formation, capacity, and retrieval of these sequences, we simulated a recurrent and a spatial spiking neural network, with heterogeneous and inhomogeneous delays respectively.

## 2.1 Neurons and network

We used the Izhikevich model [69], with a membrane-potential $v$ and a recovery variable $u$:

$$\frac{dv}{dt} = 0.04v^2 + 5v + 140 - u + I$$

$$\frac{du}{dt} = a(bv - u)$$

where $I$ is the total synaptic and external input. When $v$ reached 30 mV, the neuron fired and reset via $v \leftarrow c$ and $u \leftarrow u + d$. The parameters $(a, b, c, d)$ were $(0.02, 0.2, -65, 8)$ for excitatory regular-spiking neurons, and $(0.1, 0.2, -65, 2)$ for inhibitory fast-spiking neurons. We used Euler integration with a step size of 0.5 ms. Each network contained $N = 1000$ neurons with $N_E = 800$ excitatory and $N_I = 200$ inhibitory cells. Every neuron projected to $M = 100$ randomly chosen targets. Excitatory weights $w_{ij}$ were initialized at $w_0 = 6$ mV and were bounded within $0 \le w_{ij} \le s_{max} = 10$ mV. Inhibitory weights were fixed at $-5$ mV.

## 2.2 Spike-timing-dependent plasticity

For a connection from neuron $i$ to neuron $j$ with time delay $\tau_{ij}$, the spike-timing difference was defined as $\Delta t = t_{post} - t_{pre} - \tau_{ij}$ and the weight changes followed

$$\Delta w_{ij} = A_+ exp\left(-|\Delta t|/\tau_{STDP}\right), \qquad for\ \Delta t > 0$$
$$\Delta w_{ij} = -A_- exp\left(-|\Delta t|/\tau_{STDP}\right), \qquad for\ \Delta t < 0$$

with $A_+ = 0.10$, $A_- = 0.12$, $\tau_{STDP} = 20$ ms. Weight changes accumulated and were applied once per sec, with an additional drift of +0.01, and then clipped to $\left[0, s_{max}\right]$. Training lasted 18000 sec, where every millisecond a random neuron received an input $I_{thal} = 20$ mV.

## 2.3 Delay architectures

We simulated two networks for 18000 sec using the same plasticity rule. In the recurrent network, excitatory delays took integer values from a uniform distribution between $D_{min} = 1$ to $D_{max} = 20$ ms. Each excitatory neuron had five outgoing synapses at each delay. In the spatial network, the neurons were placed uniformly on a $4\times4$ mm cortical sheet. The connection probability decreased exponentially with the distance $l_{ij}$ and spatial decay constant $\lambda = 1.2$ mm. Excitatory delays were defined as $\tau_{ij} = \tau_{syn} + l_{ij}/v$, where the synaptic delay was $\tau_{syn} = 1$ ms and the conduction velocity was $v = 0.20$ mm/ms. We rounded delays to the nearest milliseconds and clipped them between $D_{min} = 1$ and $D_{max} = 30$. Inhibitory delays were fixed at 1 ms in both networks.

**2.4 Detecting polychronous groups**

We first selected an excitatory neuron $i$, referred to as the mother neuron. Its excitatory presynaptic neurons with weights $w_{ki} > 0.95s_{max}$ were potential anchor neurons. From these, we randomly selected up to 12 sets of three anchors $\{k_1, k_2, k_3\}$ per mother using a fixed seed. Each anchor received a brief stimulation input $I_{stim} = 1000$ mV. The stimulation time for each anchor was set to $t_{0,k} = D_{max} - \tau_{ki}$. Thus, although the three anchors fired at different times, their spikes arrived at the mother neuron at $t = D_{max}$.

We simulated the subsequent network activity for up to 120 ms, under fixed synaptic weights, without ongoing input ($I_{thal} = 0$), with an initial membrane potential $v_{init} =- 70$ mV. Letting $t_i$ denote the firing time of neuron $i$, its spike was considered a causal contributor to neuron $j$ at time $t$, when $t - 10\,ms < t_i + \tau_{ij} \leq t$. The simulation ended early if no neurons fired for $D_{max} + 1$ ms or if the cascade exceeded 100 spikes. Finally, we accepted a group only if its longest causal path met the chosen threshold ($L$≥3 or $L$≥7), every anchor contributed at least two causal edges, the total group size did not exceed 60 neurons, and its time-shifted firing pattern was unique. We excluded cases where all anchors were stimulated at the same time, as detection required asynchronous activation.

## 2.5 Reproducibility

We selected one group of 12–18 neurons and stimulated its three anchors over 10 trials. In each trial, a randomly chosen neuron received $I_{noise} = 13$ mV every millisecond, with a new random seed used for each stimulation. For each member neuron, we measured the spike-time difference compared to a noise-free trial; the mean of these differences was defined as the spike-time jitter.

## 2.6 Ablation test

We held the trained connections and weights $w_{ij}$ fixed and changed only the delays $\tau_{ij}$. In the baseline condition, delays remained unchanged. In the shuffled condition, we randomly permuted the $N_E \times M$ excitatory delay matrix (seed 7), preserving the delay distribution but changing the specific delay assigned to each synapse. In the instantaneous-delay control, all excitatory delays were set to 1 ms, the minimum value allowed by time step integration. Because a zero delay was not supported by the update scheme, removing delays corresponded to setting all excitatory delays to this minimum value. Inhibitory delays were held constant at 1 ms across all conditions.

## 2.7 Group capacity and retrieval

Network capacity was defined as the number of distinct groups detected under each delay condition, using the same detection method and candidate anchor sets. Each capacity count was reported as a percentage of the baseline value. In the recurrent network, the baseline capacity was 1137 groups for $L \geq 3$ and 31 groups for $L \geq 7$.

Retrieval tested whether the groups found at baseline could still be activated after the delays were changed. We selected 300 baseline groups, sampled evenly after sorting by causal path length, and stimulated their original anchors at their original times:

$$t_{0,k}^{base} = D_{max} - \tau_{ki}^{base}.$$

We did not adjust these stimulation times for the new delays. Each test lasted 120 ms, under fixed weights, with $v_{init} = -70$ mV, and no background input. A neuron was considered recovered if it fired within $\pm 2$ ms of its baseline spike time. A group was successfully retrieved if at least half its members were recovered. The retrieval rate was the fraction of the 300 groups meeting this criterion. While capacity quantified newly detected groups, retrieval evaluated the persistence of the original groups.

### 2.8 Population activity and energy-like score

We included the neurons belonging to any of the 30 largest groups. For each neuron $n$, the binary spike train $s_n(t)$ was smoothed with an exponential kernel

$$x_n(t) = (s_n * h)(t);\ h(t) = exp(-t/\tau_h), \qquad 0 \le t \le T_h$$

where $\tau_h = 14$ ms is the kernel time constant, and $T_h = 90$ ms is its duration. The population state $x(t)$ contains the smoothed spike trains of these neurons. For each group $k$, the reference pattern $r_k$ was the mean population state over the 90 ms window of a noise-free replay. We compared the current state with the most similar reference pattern using

$$E(x) = -log\left[\max_k cos(x, r_k)\right];\ \text{with}\ cos(x, r_k) = \frac{x \cdot r_k}{\|x\|\,\|r_k\|}.$$

Cosine similarity was clipped to $[10^{-3}, 1]$ before taking the logarithm. Each group was replayed five times with independent background noise (13 mV applied to one random neuron every millisecond). We averaged cosine similarities across trials before calculating $E(x)$. Note that this metric represents a similarity-based score, not a physical energy.

The spiking model and plasticity followed Ref. [69], and we used the simulation files and code provided on the original website (izhikevich.org) to explore polychronous groups.

## 3. Phase-Based Sequence

### 3.1 Sequence and temporal slots

Consider a sequence $S = \left(s_1, s_2, ..., s_N\right)$ of length $N$, where each item $s_i$ is drawn from a finite dictionary $V = \{v_1, v_2, ..., v_K\}$ containing $K$ possible identities. The representation separates two components of sequence information: item position is encoded by the temporal theta phase, whereas item identity is encoded by the gamma amplitude. Accordingly, a full theta cycle with frequency $f$ and period $T = 1/f$ is divided into $N$ equally spaced temporal slots. The center of slot $i$, for $i = 1, ..., N$, is defined as

$$c_i = \left(i - \frac{1}{2}\right)\frac{T}{N},$$

and the spacing between adjacent slots is $\Delta = T/N$. Thus, increasing $N$ at a fixed $T$ reduces the temporal resolution as $1/N$. The corresponding phase of slot $i$ is given by

$$\phi_i = \frac{2\pi}{N}\left(i - \frac{1}{2}\right),$$

which encodes the serial position within the sequence. Moreover, each dictionary element $v_k$ is assigned an amplitude $a_k$ within a fixed range $\left[a_{min}, a_{max}\right]$. The total dynamic amplitude range is $R = a_{max} - a_{min}$, and for uniformly spaced amplitude levels, $a_k = a_{min} + (k - 1)d$, where the spacing between neighboring identity levels is $d = R/(K - 1)$. Therefore, increasing $K$ at fixed $R$ reduces $d$ and makes neighboring items harder to identify.

### 3.2 Phase-localized encoding

Each item amplitude $a_i$ modulates a phase-localized pulse envelope $g(t)$ centered at $c_i$. The representation does not require a specific functional form for the envelope function. Instead, it is normalized at its center and required to suppress cross-talk at the other slot centers:

$$g(0) = 1, \quad g(k\Delta) = 0, \quad k \in Z,\ k \neq 0,$$

where a sufficient condition is a compact-support envelope satisfying $g(t) = 0$ for $|t| > \Delta/2$. The encoded signal over one theta cycle is given by

$$x(t) = \sum_{j=1}^{N} a_j g\left(t - c_j\right).$$

Because adjacent slot centers are separated by *Δ*, the inter-slot condition above eliminates cross-talk at the slot-center readouts under ideal phase alignment, so each item occupies a distinct phase-defined temporal slot. We have fixed $f$ = *7 Hz*. The exact envelope shape is not part of the representation; any envelope satisfying the normalization and inter-slot suppression can be used in this formulation. Reading the signal at the slot centers gives

$$r_i = x\left(c_i\right) = \sum_{j=1}^{N} a_j g\left(c_i - c_j\right),$$

or, in matrix form, $r = Ga$, where $a$ contains the encoded item amplitudes, and the mixing matrix $G$ is defined by

$$G_{ij} = g\left(c_i - c_j\right) \quad with \quad i, j = 1, 2, \ldots, N.$$

The element $G_{ij}$ quantifies the contribution of the pulse associated with item $j$ to the readout performed at slot $i$. For $i = j$, $G_{ii} = g(0) = 1$. For $i \neq j$, the slot centers differ by a nonzero integer multiple of Δ; therefore, the inter-slot condition gives $G_{ij} = 0$. Consequently, under ideal phase alignment, $G = I_N$ and $cond(G) = 1$, meaning that each slot is read independently of every other. Under ideal phase alignment, the amplitudes are recovered directly as $r_i = a_i$, without matrix inversion or deconvolution. Decoding requires one phase-locked sample per slot rather than matrix inversion or iterative calculation, and therefore has a computational complexity of $O$*(N)*.

### 3.3 Capacity of temporal order and identity

If the item at position $i$ has amplitude $a_i$ at phase $\phi_i$, the sequence representation is the pair $(\phi_i, a_i)$: phase encodes position and amplitude encodes identity. First, amplitude or readout uncertainty is modeled as additive Gaussian noise:

$$\tilde{r}_i = r_i + \varepsilon_i, \qquad \varepsilon_i \sim N\left(0, \sigma_a^2\right),$$

where $\sigma_a$ denotes the standard deviation of the amplitude noise. Thus, identity decoding depends on amplitude level relative to spacing between neighboring identity levels $d$ and noise level via $\sigma_a$, and normalized amplitude noise level can be defined as

$$\eta_{id} = \frac{\sigma_a}{d} = \frac{\sigma_a(K-1)}{R}.$$

Second, phase uncertainty is represented by a displacement $\delta_i$ of the sampling location:

$$\tilde{c}_i = c_i + \delta_i, \qquad \delta_i \sim N\left(0, \sigma_t^2\right),$$

where $\sigma_t$ denotes the standard deviation of the temporal noise introduced by random timing jitter. Thus, temporal decoding depends on displacement relative to the phase spacing between neighboring slots $\Delta$ and noise level via $\sigma_t$. The normalized temporal noise level can be defined as

$$\eta_{pos} = \sigma_t/\Delta = \sigma_t N/T.$$

Accordingly, the readout becomes

$$\tilde{r}_i = \sum_{j=1}^{N} a_j g\left(c_i + \delta_i - c_j\right) + \varepsilon_i,$$

implying that the model has two coding resolutions: identity resolution $d = R/(K - 1)$ and temporal resolution $\Delta = T/N$. Here, $R$ and $T$ are fixed, whereas $K$ and $N$ are task demands; increasing $K$ at fixed $R$ reduces the amplitude spacing $d$, whereas increasing $N$ at fixed $T$ reduces the temporal spacing $\Delta$.

The decoding performance can therefore be expressed generically as

$$P_{decode} = F\left(\eta_{id}, \eta_{pos}; g\right),$$

where the exact form depends on the pulse shape, the noise distributions, and the decoding rule. In our formulation, $K$, the number of amplitude levels, controls identity resolution and sets the level spacing $d = R/(K - 1)$ and with that $\eta_{id} = \sigma_a/d$. Independently, $N$, the number of slots, controls the temporal resolution $\Delta = T/N$, and with that $\eta_{pos} = \sigma_t/\Delta$. Thus, identity and position are limited by different but analogous resolution-noise trade-offs.

A characteristic resolution scale $K^*$ can be obtained by considering the regime in which the amplitude noise becomes comparable to the spacing between neighboring levels $d \sim \sigma_a$. Since $d = R/(K - 1)$, this gives approximately $K^* \sim R/\sigma_a$. Note that $K^*$ is a characteristic resolution scale rather than a hard bound on capacity. An operational dictionary size defined for a specified decoding accuracy will generally be smaller than $K^*$, with the difference depending on the tolerated error rate. The usable dictionary size therefore depends on the threshold in decoding accuracy.

### 3.4 Encoding–decoding mismatch

For Gaussian amplitude noise and nearest-neighbor decoding, the mean probability of correctly decoding an item is

$$P_{id} = 1 - \frac{2(K-1)}{K}\left[1 - \Phi\left(\frac{d}{2\sigma_a}\right)\right],$$

where $\Phi$ denotes the cumulative distribution function of the standard normal distribution. Using $d = R/(K - 1)$, this can equivalently be written as

$$P_{id} = 1 - \frac{2(K-1)}{K}\left[1 - \Phi\left(\frac{R}{2\sigma_a(K-1)}\right)\right].$$

Thus, increasing $K$ at fixed $R$ and $\sigma_a$ reduces the spacing between neighboring identity levels and progressively degrades the identity-decoding accuracy. An operational identity capacity can therefore be defined for a specified maximum tolerated error $\varepsilon$ as

$$K_{cap}(\varepsilon) = max\{K: P_{error}(K) \le \varepsilon\},$$

where ε is the maximum tolerated per-item identity error probability. For the temporal component, a simple slot-membership approximation can be obtained by considering a timing readout to remain within the correct temporal slot whenever $|\delta_i| < \Delta/2$.

For Gaussian temporal jitter, the corresponding slot-membership probability is approximately

$$P_{slot} = 2\Phi\left(\frac{\Delta}{2\sigma_t}\right) - 1,$$

using $\Delta = T/N$, this can equivalently be written as

$$P_{slot} = 2\Phi\left(\frac{T}{2N\sigma_t}\right) - 1.$$

Therefore, increasing $N$ at fixed $T$ reduces $\Delta$ and makes the representation increasingly sensitive to temporal jitter. The exact effect of timing jitter on decoding additionally depends on the chosen pulse envelope *g*. If amplitude and temporal errors are treated approximately as independent, the probability of correctly decoding an individual item can be approximated as $P_{item} \approx P_{id}P_{slot}$. Under ideal phase alignment and no temporal jitter, $\sigma_t = 0$, we have $P_{slot} = 1$, and for independent slot error $P_{item} \approx P_{id}(K)$, therefore increasing $N$ does not reduce per-item decoding accuracy. An $N$-dependent reduction in per-item accuracy appears only when temporal uncertainty is introduced because the slot width $\Delta = T/N$ decreases as $N$ increases. Importantly, the probability of recovering the complete $N$-item sequence is different from per-item accuracy. If item errors are approximately independent across slots $P_{seq} = \left(P_{id}(K)\right)^N$, and the information per cycle scales as $N\, log_2 K$.